\documentclass[runningheads]{llncs}

\usepackage{eccv}

\usepackage{eccvabbrv}

\usepackage{graphicx}
\usepackage{booktabs}
\usepackage{multirow}
\usepackage{wrapfig}
\usepackage[table]{xcolor}
\usepackage{pifont} % for checkmark
\newcommand{\cmark}{\ding{51}}
\newcommand{\xmark}{\ding{55}}
\definecolor{customOrange}{HTML}{EA8121}
\definecolor{customGreen}{HTML}{2FA85F}
\definecolor{customPurple}{HTML}{9D6CA6}

\definecolor{suppGreen}{RGB}{79, 151, 79}
\usepackage[accsupp]{axessibility}  % Improves PDF readability for those with disabilities.

\usepackage{hyperref}

\usepackage{orcidlink}

\begin{document}

% % ---------------------------------------------------------------
% % TODO REVIEW: Replace with your title
\title{$C^3$ASD: Multi-Level Consistency-Driven Representation Learning for Robust Active Speaker Detection} 
% %Robust Audio-Visual Active Speaker Detection via Multi-Level Consistency Learning

% % TODO REVIEW: If the paper title is too long for the running head, you can set
% % an abbreviated paper title here. If not, comment out.
\titlerunning{$C^3$ASD: Multi-Level Consistency-Driven Representation Learning}
% % TODO FINAL: Replace with your author list. 
% % Include the authors' OCRID for the camera-ready version, if at all possible.
% Second Author\inst{2,3}\orcidlink{1111-2222-3333-4444} \and
% Third Author\inst{3}\orcidlink{2222--3333-4444-5555}}

\author{Jin Hong\inst{1}\textsuperscript{*}\orcidlink{0009-0002-7101-8040} \and Jisoo Park\inst{1}\textsuperscript{*}\orcidlink{0009-0004-0885-9538} \and Junseok Kwon\inst{1}\orcidlink{0000-0001-9526-7549}}

% % TODO FINAL: Replace with an abbreviated list of authors.
\authorrunning{J. Hong et al.}

% % TODO FINAL: Replace with your institution list.
% Springer Heidelberg, Tiergartenstr.~17, 69121 Heidelberg, Germany

%\institute{Chung-Ang University, Seoul, Republic of Korea\\ \email{\{jindl465,susiehome,jskwon\}@cau.ac.kr} }

\institute{Chung-Ang University, Seoul, Republic of Korea\\ \email{\{jindl465,susiehome,jskwon\}@cau.ac.kr}\\ \textsuperscript{*}Equal contribution.}

\maketitle

\begin{center}
\refstepcounter{figure}
\includegraphics[width=0.85\linewidth]{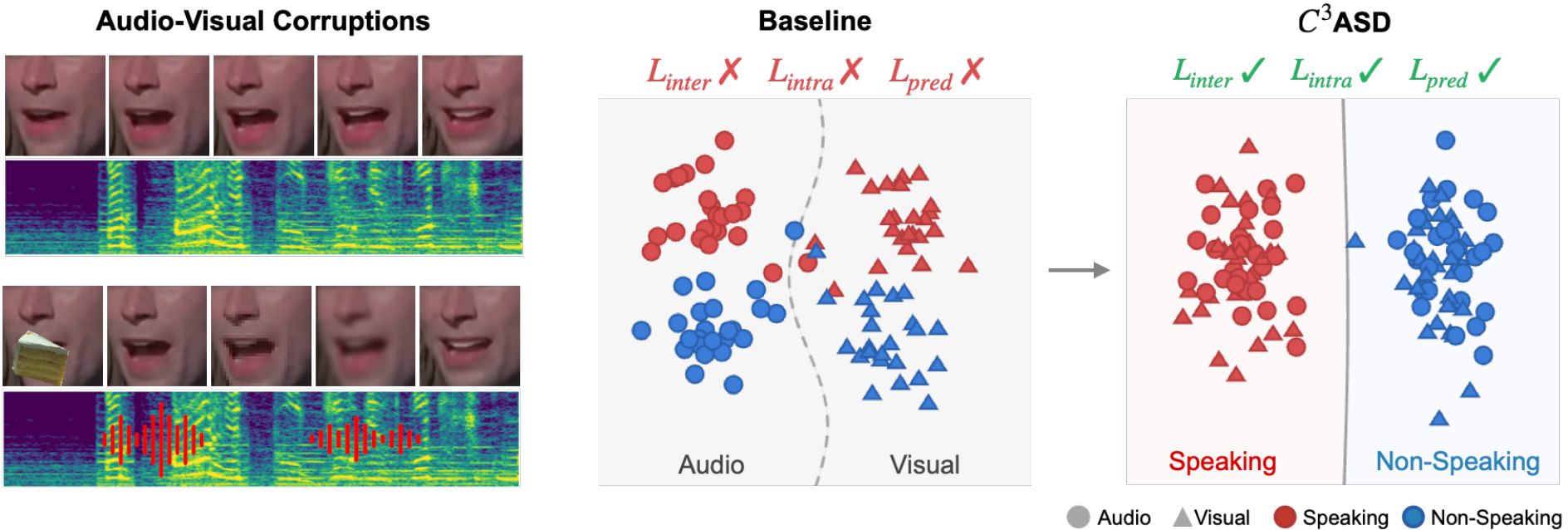}
\parbox{0.85\linewidth}{\small
\textbf{Fig.~\thefigure.}
\textbf{Multi-Level Consistency for Robust ASD.} In real-world active speaker detection, audio and visual streams are often corrupted by noise and occlusion. Existing models learn modality-clustered representations fragile under such degradations. In contrast, our multi-level consistency framework produces class-aligned, modality-invariant embeddings robust across diverse corruption scenarios. \label{fig:teaser}}
\end{center}
\vspace{3mm}

\begin{abstract}
\vspace{-7mm}
  Active Speaker Detection determines whether a visible person in a video is speaking at each moment. While recent audio–visual fusion methods perform well on clean data, they degrade under real-world corruptions such as background noise, occlusion, or simultaneous modality degradation. We attribute this limitation to the absence of explicit consistency constraints that promote robust, semantically aligned representations across modalities. Without such guidance, models tend to learn fragile modality-specific shortcuts that fail under corrupted conditions. We propose $C^3$ASD, a multi-level consistency-driven framework with three complementary constraints: embedding-level inter-modality consistency aligns audio-visual representations during speech; sequence-level intra-modality consistency separates speaking and non-speaking clusters via track-aware contrastive learning; and prediction-level consistency stabilizes fusion through knowledge distillation. Extensive experiments demonstrate significant improvements under diverse audio, visual and joint corruptions, while maintaining competitive performance on clean data.
\end{abstract}
\section{Introduction}
\label{sec:intro}

Active Speaker Detection (ASD) determines whether a visible person in a video is speaking at each moment, serving as a fundamental task in video understanding with applications in video conferencing~\cite{videoconf}, human–robot interaction~\cite{hci,hci_robot,hci_robot2,hci_robot3}, and multimedia retrieval~\cite{retreival}. Owing to its inherently multimodal nature, ASD relies on both auditory speech signals and visual facial movements, motivating the development of audio–visual fusion methods that exploit complementary information from the two modalities. Recent advances in audio–visual ASD have achieved strong performance using sophisticated fusion architectures, such as temporal convolutional networks, transformers, and attention mechanisms~\cite{talknet, lightasd, spell, loconet, unicon}.
These methods typically extract modality-specific features and aggregate them through various fusion strategies.
Despite their success on clean data, they often treat audio and visual streams as separate sources to be combined, without explicitly modeling how these modalities should relate to and constrain each other. This oversight becomes particularly problematic in real-world scenarios where one or both modalities may be corrupted by noise, occlusion, or other distortions.

In real-world settings, modality corruption is inevitable. Audio may be contaminated by background noise, music, or overlapping speech~\cite{musan, demand}, while visual inputs can degrade due to motion blur, occlusion, low resolution, or rapid head movement. More challenging are scenarios where both modalities are simultaneously unreliable, such as unconstrained videos in the wild~\cite{wasd}.
Under such conditions, existing ASD models often experience significant performance degradation~\cite{cav2vec, rasd}, indicating that current approaches fail to robustly and complementarily leverage audio and visual information.

We attribute this limitation to how current models learn audio-visual representations. As shown in~\cref{fig:teaser}, existing models learn modality-clustered representations that are fragile under real-world corruptions, whereas our approach produces class-aligned, modality-invariant embeddings robust across diverse degradation scenarios. Although prior methods leverage both modalities, they typically lack explicit mechanisms to enforce consistent representations, instead fusing audio and visual features via simple operations such as summation or concatenation optimized primarily with classification objectives~\cite{lightasd, talknet}. Without additional structural constraints, models learn modality-specific shortcuts that fail to capture modality-invariant speaking characteristics and generalize poorly when input conditions deviate from training distributions.

To address this limitation, we propose $C^3$ASD, a \emph{multi-level consistency-driven representation learning framework} built upon three complementary consistency constraints for robust audio-visual ASD. Our central idea is to explicitly regularize representation learning so that speaking-related information is encoded in a stable and modality-consistent manner. Specifically, we introduce three complementary consistency constraints that impose structural relationships within and across modalities.
\emph{(1) Inter-modality consistency} aligns audio and visual embeddings when a person is actively speaking, as both modalities reflect the same underlying speech production process~\cite{clip, xdc, syncnet}. Importantly, this constraint is applied only during actual speech; enforcing audio-visual similarity during silence, when both modalities capture only background noise, provides no useful learning signal and may introduce spurious correlations.
\emph{(2) Intra-modality consistency} encourages each modality to form well-separated embedding clusters for speaking and non-speaking states~\cite{contrastive, vicreg}. While inter-modality alignment promotes cross-modal coherence, it does not constrain the internal structure of individual modalities, leaving embeddings vulnerable to small perturbations near the decision boundary. By pulling same-label embeddings closer and pushing different-label embeddings apart within each modality, intra-modality consistency enlarges the margin between clusters, reducing the risk that corruption-induced shifts alter the final prediction.
\emph{(3) Prediction-level consistency} stabilizes multimodal fusion by enforcing agreement between multimodal and unimodal predictions via mean squared error, inspired by knowledge distillation~\cite{distilling, meanteacher}. This keeps unimodal branches calibrated to the more reliable multimodal output, preventing the final decision from being dominated by a corrupted modality.

These three consistency constraints collectively guide the model to learn representations that capture the underlying speaking state rather than modality-specific artifacts. By explicitly regularizing the representation learning, our approach improves robustness under audio, visual, and joint corruptions while maintaining competitive performance on clean data.
The method is lightweight, requires no additional annotations, and can be seamlessly integrated into existing ASD architectures with minimal modification. Our key contributions are:

\noindent$\blacktriangleright$ We reveal that the absence of explicit multi-level modality consistency constraints fundamentally limits the robustness of existing audio–visual ASD models under real-world corruptions.

\noindent$\blacktriangleright$ We design three consistency constraints to explicitly regularize audio-visual representation learning: embedding-level inter-modality consistency with speaking-aware alignment, intra-modality consistency via sequence-level regularization, and prediction-level consistency through knowledge distillation.

\noindent$\blacktriangleright$  Extensive experiments demonstrate that our method significantly improves robustness under diverse audio, visual, and joint corruptions while preserving competitive performance on clean data.

%Through ablation studies, we reveal the complementary roles of different consistency types: prediction-level consistency improves clean performance, while inter-modality consistency provides critical cross-modal compensation under corruption.
\section{Related Works}
\label{sec:related_works}

\noindent\textbf{Audio-Visual Active Speaker Detection.} 
%Audio-Visual Active Speaker Detection + 이모델이 robustness에 한계있음을 언급
ASD aims to determine whether each visible person in a video is speaking at a given moment. AVA-ActiveSpeaker dataset~\cite{ava} has become the standard benchmark for evaluating ASD methods, where recent approaches~\cite{adenet,asc,asdtrans,easee,loconet,talknet,lightasd,lrasd,talknce,synctalknet} have achieved strong performance under controlled settings.
Early deep learning methods such as TalkNet~\cite{talknet} introduced temporal convolutional architectures with multi-scale feature aggregation to model long-range audio–visual correspondence. Subsequent works explored richer contextual modeling: SPELL~\cite{spell} leveraged spatial–temporal graph learning to capture inter-speaker relationships, UniCon~\cite{unicon} proposed unified context networks integrating spatial, relational, and temporal cues, and LoCoNet~\cite{loconet} modeled both long-term intra-speaker dependencies and short-term inter-speaker interactions. EASEE~\cite{easee} further advanced the pipeline by jointly optimizing face detection and ASD in an end-to-end framework.
Parallel efforts have focused on computational efficiency. Light-ASD~\cite{lightasd} demonstrated that lightweight architectures can maintain competitive performance with significantly reduced complexity, and LR-ASD~\cite{lrasd} improved efficiency and long-range dependency modeling.

Despite their strong benchmark performance, these methods are largely optimized for clean conditions where both modalities are reliable. Their fusion strategies typically lack explicit mechanisms to enforce robust and consistent cross-modal representations. This limitation becomes critical in real-world scenarios, where audio and visual signals are frequently corrupted.

\noindent\textbf{Robust Active Speaker Detection.}
%RobustASD(corruption을 다루는 ASD 모델) + 이 모델의 여전한 한계있음
%UniTalk: Towards Universal Active Speaker Detection in Real World Scenarios 논문&데이터셋 사용가능할지 확인필요
To address real-world deployment challenges, several works have investigated ASD under adverse conditions, including background noise, non-speech interference, occlusions, and complex scenes. A notable direction explicitly formulates robust ASD in noisy environments~\cite{rasd}, highlighting that environmental sounds can corrupt audio representations and degrade performance. One solution introduces audio–visual speech separation as auxiliary guidance to learn noise-resistant audio features. 
In parallel, methods such as UniCon~\cite{unicon} improved robustness through enhanced contextual modeling and stronger audio–visual fusion under diverse conditions. Other approaches emphasized end-to-end embedding learning and context aggregation to enhance reliability~\cite{easee}, reducing dependence on multi-stage pipelines. Graph-based formulations, such as SPELL~\cite{spell}, modeled ASD as node classification over multimodal temporal graphs to better capture long-range and inter-person dependencies.

Despite these advances, robustness remains limited under modality-specific corruption (audio-only or visual-only) and, more critically, under joint corruption where both streams are unreliable. Most existing methods focus primarily on architectural improvements or fusion strategies, but lack explicit training signals that enforce stability and cross-modal compatibility in the learned representations and predictions. This gap motivates a representation-level regularization approach rather than further architectural complexity.

\noindent\textbf{Consistency-Based Representation Learning.}
%Consistency-Based Representation Learning + 이게 ASD에 왜 맞을지
Consistency-based regularization is a widely adopted principle for learning stable representations, with contrastive learning emerging as an effective mechanism for enforcing invariance in representation space. Foundational works such as SimCLR~\cite{simclr} and MoCo~\cite{moco} demonstrate that instance discrimination and invariance objectives substantially improve representation quality and generalization. Beyond contrastive approaches, negative-pair-free methods such as BYOL~\cite{byol} and SimSiam~\cite{simsiam} further demonstrate that consistency alone, without explicit negatives, suffices for robust representation learning. In multimodal learning, consistency objectives play an equally central role: cross-modal alignment methods such as CLIP~\cite{clip} show that explicitly matching representations across modalities yields robust and transferable features. Similarly, audio–visual self-supervised approaches (\eg AV-HuBERT~\cite{avhubert}, XDC~\cite{xdc}) learn modality-aligned representations without dense supervision, further validating the importance of cross-modal consistency.
In the context of ASD, recent work has begun to incorporate contrastive objectives to enhance representation quality~\cite{talknce}, indicating that performance gains extend beyond architectural design. However, these approaches are primarily evaluated under standard benchmark conditions and do not explicitly address representation instability under real-world modality corruption. This issue is particularly critical for ASD, where audio and visual streams share the same supervision target but vary in reliability over time.

\noindent\textbf{Novelty of Our Work.}
Existing ASD methods address robustness primarily through architectural improvements or fusion strategies, while consistency-based representation learning methods have largely been validated under clean, uncorrupted conditions. Our work bridges these two gaps: rather than modifying the backbone architecture, we introduce a multi-level consistency framework that brings representation-level regularization explicitly into the ASD training objective.
%We formulate consistency not as an auxiliary training signal, but as a structured multi-level framework that explicitly regularizes cross-modal alignment, intra-modality structure, and prediction agreement to improve robustness under modality-specific and joint corruption.
\section{Proposed Method}

\begin{figure*}[t]
    \begin{minipage}[t]{\linewidth}
	\centering
  \includegraphics[width=1.0\linewidth]{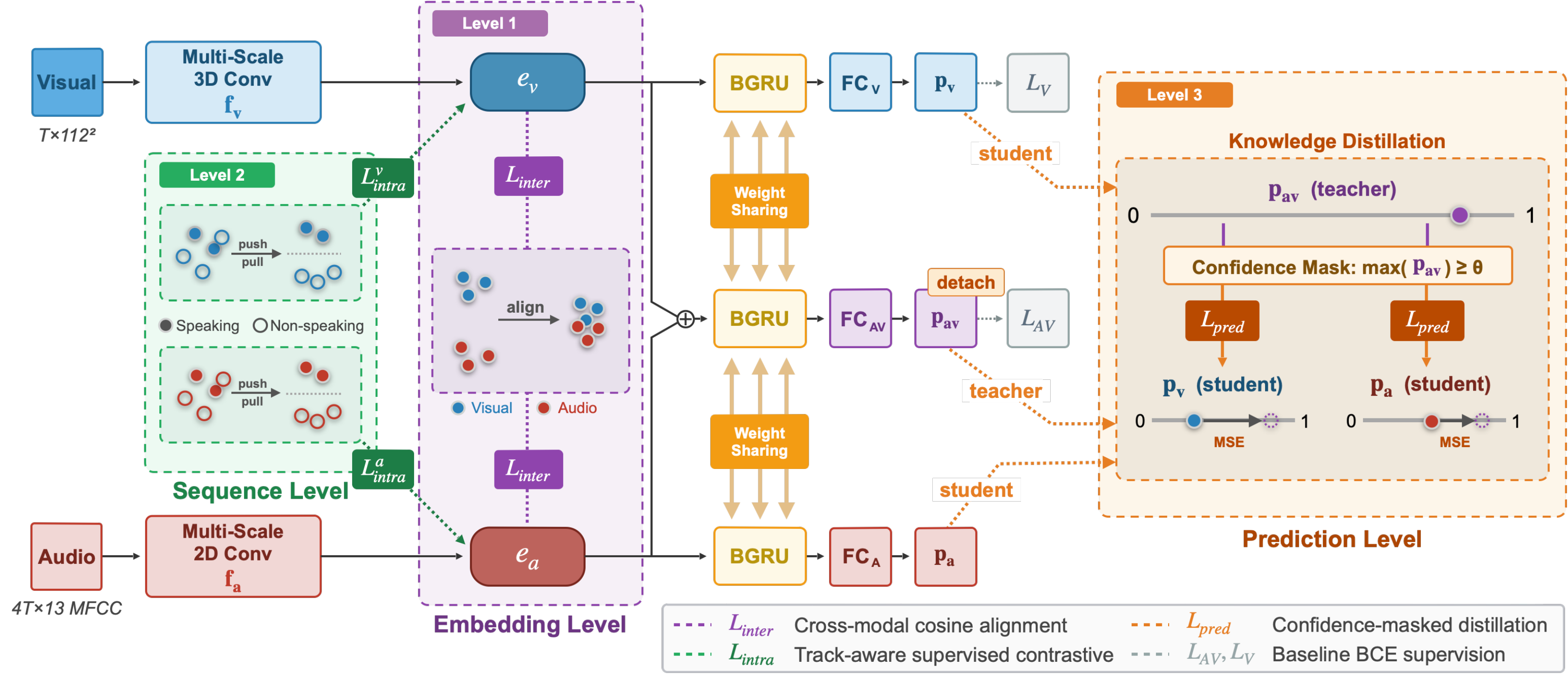}
    \end{minipage}
  \caption{\textbf{Overview of the proposed $C^3$ASD framework.} The model takes visual face crops and audio MFCC (Mel-Frequency Cepstral Coefficients) features as input and processes them through modality-specific encoders ($f_v$, $f_a$). We introduce three complementary consistency constraints at multiple levels: (1) Embedding-level inter-modality consistency (\textcolor{customPurple}{$\mathcal{L}_{\text{inter}}$}) aligns audio and visual embeddings via cosine similarity during active speech; (2) Sequence-level intra-modality consistency (\textcolor{customGreen}{$\mathcal{L}_{\text{intra}}$}) applies track-aware supervised contrastive learning within each modality to separate speaking and non-speaking clusters; and (3) Prediction-level consistency (\textcolor{customOrange}{$\mathcal{L}_{\text{pred}}$}) distills knowledge from the confidence-masked audio--visual prediction $p_{av}$ (teacher) to unimodal predictions $p_v$ and $p_a$ (students) via MSE on speaking probabilities. All three BGRUs share weights, and the framework introduces no additional learnable parameters beyond a lightweight classification head.}
  \label{fig:overview}
  \vspace{-7mm}
\end{figure*}

We address robust ASD, which determines whether a visible person in a video is speaking at each time step. Building upon Light-ASD~\cite{lightasd}, we introduce \emph{multi-level consistency regularization} to enhance robustness against real-world audio-visual corruptions without requiring corrupted training data. \cref{fig:overview} illustrates the overall framework.

\subsection{Baseline Architecture}

Our backbone follows Light-ASD~\cite{lightasd}, a lightweight two-stream architecture composed of a visual encoder $f_v$, an audio encoder $f_a$, and a Bidirectional GRU (BGRU)~\cite{gru} classifier $g$.
The visual encoder $f_v$ takes a sequence of $T$ grayscale face crops ($112 \times 112$) detected by S3FD~\cite{s3fd} and produces frame-level embeddings $\mathbf{e}_v \in \mathbb{R}^{B \times T \times 128}$ using three multi-scale 3D convolutional blocks with factorized spatiotemporal kernels (sizes 3 and 5), intermediate max-pooling, and adaptive spatial pooling. 
The audio encoder $f_a$ processes $4T$ MFCC frames (13-dimensional, extracted at $4\times$ the video frame rate) through a similar multi-scale 2D convolutional architecture, yielding $\mathbf{e}_a \in \mathbb{R}^{B \times T \times 128}$.
Audio–visual fusion is performed via element-wise addition, $\mathbf{e}_{av} = \mathbf{e}_a + \mathbf{e}_v$, followed by a BGRU with GELU activations for temporal modeling. Each prediction head maps the 128-dimensional features to binary logits through a fully connected layer with temperature-scaled softmax, where the temperature is scheduled as $r = 1.3 - 0.02 \times (e-1)$ at epoch $e$.

The baseline training objective combines audio–visual and visual-only losses:
\begin{equation}
\mathcal{L}_{\text{base}} = \mathcal{L}_{\text{AV}} + 0.5 \, \mathcal{L}_{\text{V}},
\label{eq:base_loss}
\end{equation}
where each term $\mathcal{L}_{s \in \{\text{AV}, \text{V}\}}$ denotes the standard binary cross-entropy loss between the predicted speaking probability and the ground-truth label.

\subsection{Multi-Level Consistency Regularization}
Our three complementary consistency losses (\ie inter-modality, intra-modality, prediction-level consistency) operate at both embedding and prediction levels. They require no external data or architectural modifications and are designed to strengthen the internal coherence of learned representations, enabling the model to degrade gracefully when either modality is corrupted at test time.

\noindent\textbf{Embedding-Level Inter-Modality Consistency.}
The audio and visual streams observe the same underlying speaking event through different sensor modalities. We align their latent representations to encourage each encoder to capture modality-invariant speaking cues, introducing redundancy that improves robustness: when one modality is degraded at test time, the other can compensate through the shared representational structure.
Concretely, we maximize the cosine similarity between corresponding frame-level audio and visual embeddings. For the $i$-th speaking frame ($y_i = 1$) in a batch, letting $\mathcal{S} = \{i \mid y_i = 1\}$, the inter-modality loss is defined as,
\begin{equation}
    \mathcal{L}_{\text{inter}} = \frac{1}{|\mathcal{S}|} \sum_{i \in \mathcal{S}} \left(1 - {\cos(\mathbf{e}_a^{(i)}, \mathbf{e}_v^{(i)})}\right) =\frac{1}{|\mathcal{S}|} \sum_{i \in \mathcal{S}} \left(1 - \frac{\mathbf{e}_a^{(i)} \cdot \mathbf{e}_v^{(i)}}{\|\mathbf{e}_a^{(i)}\| \cdot \|\mathbf{e}_v^{(i)}\|}\right).
    \label{eq:inter_loss}
\end{equation}

This loss operates on the raw encoder outputs \textit{before} fusion, directly shaping the geometry of each modality’s embedding space. Unlike contrastive objectives that rely on explicit negative pairs, the cosine similarity term serves as a soft alignment constraint without collapsing representations into a trivial solution, consistent with findings in negative-pair-free learning~\cite{byol,simsiam}. Each modality preserves its discriminative structure through the supervised losses $\mathcal{L}_{\text{AV}}$ and $\mathcal{L}_{\text{V}}$ in \cref{eq:base_loss}, while $\mathcal{L}_{\text{inter}}$ enforces cross-modal coherence.
This alignment is crucial for robustness. When audio is corrupted by background noise, the audio embedding may drift from its clean representation; however, a well-aligned visual embedding can anchor the fused representation $\mathbf{e}_{av} = \mathbf{e}_a + \mathbf{e}_v$ toward the correct decision boundary.

\noindent\textbf{Sequence-Level Intra-Modality Consistency.}
Inter-modality alignment alone is insufficient to guarantee robust representations, as it does not constrain the internal geometry of each modality.
We observe that embeddings with clear class separation, where speaking frames cluster together and non-speaking frames form a distinct cluster, are more robust to perturbations. When the margin between clusters is large, corrupted inputs are less likely to cross the decision boundary.
To enforce this structure, we apply a supervised contrastive loss~\cite{contrastive} independently to each modality. 
Given $\ell_2$-normalized frame-level embeddings $\bar{\mathbf{e}}_i = \mathbf{e}_i / \|\mathbf{e}_i\|$ and binary speaking labels $y_i \in \{0, 1\}$, the intra-modality loss is defined as:
\begin{equation}
    \mathcal{L}_{\text{intra}} = -\frac{1}{N} \sum_{i=1}^{N} \frac{1}{|P(i)|} \sum_{j \in P(i)} \log \frac{\exp\left(\bar{\mathbf{e}}_i \cdot \bar{\mathbf{e}}_j \;/\; \tau\right)}{\displaystyle\sum_{k \in \mathcal{N}(i)} \exp\left(\bar{\mathbf{e}}_i \cdot \bar{\mathbf{e}}_k \;/\; \tau\right)},
    \label{eq:intra_loss}
\end{equation}
where $\tau = 0.07$ is a temperature hyperparameter.

A critical design choice concerns the definition of the positive set $P(i)$ and the negative set $\mathcal{N}(i)$ in \cref{eq:intra_loss}.
In ASD, each training batch typically contains multiple face tracks from different speakers. Naively treating all same-label frames as positives would incorrectly pull together embeddings of \textit{different} speakers who happen to be speaking simultaneously, introducing noise into the contrastive objective.
To address this, we introduce a \textit{track-aware constraint}. Let $g_i$ denote the track identity of frame $i$. The positive set is defined as
\[
P(i) = \{ j \neq i \mid y_j = y_i,\; g_j = g_i \},
\]
\ie frames with the same speaking label \textit{and} the same track identity. The denominator sums over
\[
\mathcal{N}(i) = \{ k \neq i \mid g_k = g_i \},
\]
which includes all other frames within the same track. 
This design confines contrastive learning to each speaker’s temporal context, producing semantically coherent clusters while avoiding cross-speaker interference.
The loss is applied independently to audio and visual embeddings, yielding $\mathcal{L}_{\text{intra}}^{a}$ and $\mathcal{L}_{\text{intra}}^{v}$ with separate weighting coefficients, as the two modalities may require different regularization strengths. Empirically, stronger intra-modality regularization on the visual stream proves beneficial, since visual corruptions (\eg occlusion and blur) tend to induce larger embedding shifts than typical audio noise.

\noindent\textbf{Prediction-Level Consistency.}
The embedding-level consistency losses structure the feature space but do not explicitly regulate the \textit{output predictions} of individual modality streams. To further stabilize multimodal decision-making, we introduce a prediction-level consistency loss that transfers knowledge from the stronger audio–visual stream to the unimodal streams, encouraging reliable predictions even when a single modality is used.
Specifically, the audio–visual prediction $\mathbf{p}_{av}$ serves as a teacher (with gradients detached), while the audio-only $\mathbf{p}_a$ and visual-only $\mathbf{p}_v$ predictions act as students:
\begin{equation}
    \mathcal{L}_{\text{pred}} = \frac{1}{|\mathcal{M}|} \sum_{i \in \mathcal{M}} \left[\left(p_a^{(i)} - p_{av}^{(i)}\right)^2 + \left(p_v^{(i)} - p_{av}^{(i)}\right)^2\right],
    \label{eq:pred_loss}
\end{equation}
where $p_{av}^{(i)}$, $p_a^{(i)}$, and $p_v^{(i)}$ denote the speaking probability (softmax output for the positive class) of each stream at frame $i$.
The audio-only prediction is obtained by passing the audio embedding $\mathbf{e}_a$ through the shared BGRU followed by a dedicated audio classification head.

In \cref{eq:pred_loss}, teacher predictions may vary in reliability. To mitigate noisy supervision, inspired by confidence-based filtering in pseudo-label learning~\cite{pseudolabel} and mean teacher methods~\cite{meanteacher}, we apply a confidence mask
\[
\mathcal{M} = \{ i \mid \max\left(\mathbf{p}_{av}^{(i)}\right) \geq \theta \},
\]
with threshold $\theta = 0.7$, to select only high-confidence audio–visual predictions as supervision targets.
This filtering prevents unimodal streams from learning from noisy or ambiguous teacher outputs, which are common in early training when the model is poorly calibrated. As training progresses and the teacher becomes more confident, the effective supervision set $\mathcal{M}$ naturally expands, resulting in an implicit curriculum learning effect.

This loss provides a distinct and complementary signal to the embedding-level objectives. 
While $\mathcal{L}_{\text{inter}}$ aligns cross-modal representations and $\mathcal{L}_{\text{intra}}$ structures the geometry within each modality, $\mathcal{L}_{\text{pred}}$ directly penalizes prediction disagreement, ensuring that improvements in representation space translate into consistent output behavior.
At test time, even when one modality is severely corrupted, the regularized unimodal streams yield more stable predictions, preventing the fused output from being dominated by a single unreliable modality.

\subsection{Total Training Objective}
The overall training objective combines the baseline loss in \cref{eq:base_loss} with the three consistency regularizers in \cref{eq:inter_loss}, \cref{eq:intra_loss}, and \cref{eq:pred_loss}:
\begin{equation}
    \mathcal{L}_{\text{total}} = \underbrace{\mathcal{L}_{\text{AV}} + 0.5 \cdot \mathcal{L}_{\text{V}}}_{\text{baseline}} + \lambda_1 \mathcal{L}_{\text{inter}} + \lambda_2 \mathcal{L}_{\text{intra}}^{a} + \lambda_3 \mathcal{L}_{\text{intra}}^{v} + \lambda_4 \mathcal{L}_{\text{pred}},
    \label{eq:total_loss}
\end{equation}
where $\lambda_1, \lambda_2, \lambda_3,$ and $\lambda_4$ balance the contributions of each term.
All parameters are optimized end-to-end using Adam~\cite{adam} with an initial learning rate of $10^{-3}$ and exponential decay ($\gamma = 0.95$ per epoch) for 30 epochs.
Importantly, the proposed consistency losses introduce no additional learnable parameters, except for the lightweight audio-only classification head (a single $128 \rightarrow 2$ linear layer), thereby preserving the efficiency of the baseline architecture.

% \begin{equation}
%     \mathcal{L}_{\text{total}} =
%     \underbrace{
%         \mathcal{L}_{\text{AV}} + 0.5 \mathcal{L}_{\text{V}}
%     }_{\substack{\text{baseline}\\\text{supervision}}}
%     +
%     \underbrace{
%         \lambda_1 \mathcal{L}_{\text{inter}}
%     }_{\substack{\text{Level 1}\\\text{(align)}}}
%     +
%     \underbrace{
%         \lambda_2 \mathcal{L}_{\text{intra}}^{a}
%         + \lambda_3 \mathcal{L}_{\text{intra}}^{v}
%     }_{\substack{\text{Level 2}\\\text{(compact)}}}
%     +
%     \underbrace{
%         \lambda_4 \mathcal{L}_{\text{pred}}
%     }_{\substack{\text{Level 3}\\\text{(stabilize)}}},
%     \label{eq:total_loss}
% \end{equation}
\section{Experiments}
\vspace{-2mm}
%\subsection{Datasets}

\noindent\textbf{In-Domain Benchmark.}
We used the AVA-ActiveSpeaker dataset~\cite{ava} for training and in-domain evaluation, consisting of 262 Hollywood movie clips with frame-level speaking activity labels. The training and validation splits contain approximately 5.3M and 0.75M labeled face tracks, respectively. Performance is measured using mean Average Precision (mAP) following the official protocol.

\noindent\textbf{In-the-Wild Benchmark.}
We used the WASD dataset~\cite{wasd} to evaluate generalization beyond in-domain performance. WASD contains videos from diverse real-world scenarios with challenging conditions such as varying lighting, camera motion, and background noise, providing a complementary evaluation to the relatively controlled AVA setting.

\noindent\textbf{Corruption Benchmark.}
To evaluate robustness under real-world degradation, we constructed a corruption benchmark applied exclusively at test time with no corrupted training samples, following CAV2Vec~\cite{cav2vec}. For audio corruptions, we used MUSAN~\cite{musan} (babble, music, and ambient noise at SNR $\in [-10, 10]$ dB) and DEMAND~\cite{demand} (eight real-world noise environments). For visual corruptions, we applied object occlusion (COCO patches~\cite{coco} + Gaussian noise) and pixelation to $112\times112$ face crops. 

\noindent\textbf{Comparison Models.}
We compared against methods trained from scratch without pre-trained backbone features, matching our setup (gray rows in \cref{tab:ava}). We selected TalkNet~\cite{talknet}, ADENet~\cite{adenet}, and Light-ASD~\cite{lightasd}, which provide public implementations. Other methods are excluded due to unavailable source code~\cite{synctalknet, asdtrans} or architectural overlap~\cite{lrasd} with included baselines.

\vspace{-3mm}
\subsection{Implementation Details}
\vspace{-2mm}
We trained all models on the AVA-ActiveSpeaker training set for 30 epochs using the Adam optimizer~\cite{adam} with an initial learning rate of $10^{-3}$ and a step decay schedule ($\gamma = 0.95$ per epoch). The dynamic batch size was set to 1,000 frames. Audio features consisted of 13-dimensional Mel-Frequency Cepstral Coefficients (MFCCs) extracted from 16kHz audio, with window and hop sizes aligned to the video frame rate. Visual inputs were grayscale face crops resized to $112 \times 112$ pixels.
For the proposed consistency regularization, the loss weights were set to $\lambda_1 = \lambda_4 = 0.01$ and $\lambda_2 = \lambda_3 = 0.001$. The intra-modality contrastive temperature was $\tau = 0.07$, and the prediction-level confidence threshold was set to $\theta = 0.7$. These hyperparameters were selected based on validation performance and kept fixed across all experiments. All models were trained using PyTorch~\cite{pytorch} on a single NVIDIA A6000 GPU.

% % --------------- indomain test -------------------%
\begin{table}[t]
\centering
\caption{\textbf{(In-domain) Comparison of recent ASD methods on AVA.}
We report parameters, FLOPs, and mAP.
Methods trained from scratch with end-to-end learning (our experimental comparison set) are highlighted in gray. \textit{*For Light-ASD, we report the reproduced results.}}
\vspace{-3mm}
\label{tab:ava}
\setlength{\tabcolsep}{7pt}
\renewcommand{\arraystretch}{0.7}
\scriptsize
\begin{tabular}{lccc|ccc}
\toprule
Method 
& Single 
& Pre-training 
& E2E
& Params(M) 
& FLOPs(G) 
& mAP(\%) \\
\midrule
ASC~\cite{asc}           & \xmark & \cmark & \xmark & 23.5  & 1.8  & 87.1 \\
MAAS~\cite{maas}         & \xmark & \cmark & \xmark & 22.5  & 2.8  & 88.8 \\
UniCon~\cite{unicon}       & \xmark & \cmark & \xmark & $>$22.4 & $>$1.8 & 92.2 \\
ASDNet~\cite{asdnet}     & \xmark & \cmark & \xmark & 51.3  & 14.9 & 93.5 \\
EASEE-50~\cite{easee}    & \xmark & \cmark & \cmark & $>$74.7 & $>$65.5 & 94.1 \\
SPELL~\cite{spell}           & \xmark & \cmark & \xmark & 22.5 & 2.6 & 94.2 \\
SPELL+~\cite{spellplus}          & \xmark & \cmark & \xmark & 47.3 & 5.4 & 94.9 \\
LoCoNet~\cite{loconet}      & \xmark & \cmark & \cmark & 34.3 & 4.9 & 95.2 \\
TalkNCE~\cite{talknce}      & \xmark & \cmark & \cmark & 34.3 & 4.9 & \textbf{95.5} \\
\midrule
\rowcolor{gray!10} TalkNet~\cite{talknet}         & \cmark & \xmark & \cmark & 15.7 & 1.5 & 92.3 \\
\rowcolor{gray!10} Sync-TalkNet~\cite{synctalknet} & \cmark & \xmark & \cmark & 15.7 & 1.5 & 89.8 \\
\rowcolor{gray!10} ASD-Transformer~\cite{asdtrans} & \cmark & \xmark & \cmark & $>$13.9 & $>$1.5 & 93.0 \\
\rowcolor{gray!10} ADENet~\cite{adenet}         & \cmark & \xmark & \cmark & 33.2 & 22.7 & 93.2 \\
\rowcolor{gray!10} Light-ASD*~\cite{lightasd}   & \cmark & \xmark & \cmark & 1.02 & 0.62 & 93.6 \\
\rowcolor{gray!10} LR-ASD~\cite{lrasd}          & \cmark & \xmark & \cmark & \textbf{0.84} & \textbf{0.51} & 94.5 \\
\rowcolor{gray!10} \textbf{Ours}                         & \cmark & \xmark & \cmark & 1.02 & 0.62 & 93.8 \\
\bottomrule
\end{tabular}
\vspace{-5mm}
\end{table}

\subsection{In-Domain Results on AVA-ActiveSpeaker}
\cref{tab:ava} presents a comparison between our method and state-of-the-art ASD approaches on the AVA-ActiveSpeaker validation set. For each method, we report the number of parameters, computational cost (FLOPs), and mAP. The methods are grouped into two categories: those leveraging large pretrained backbones and those trained from scratch. Among lightweight models trained from scratch, our method achieves 93.8\% mAP with only 1.02M parameters and 0.62G FLOPs, comparable to Light-ASD~\cite{lightasd} without additional architectural complexity. Notably, our method outperforms heavier models such as TalkNet~\cite{talknet} and ADENet~\cite{adenet} at a significantly smaller parameter count, and all consistency-regularized variants maintain in-domain mAP within 0.3\% of the baseline, confirming that the proposed losses preserve both efficiency and accuracy.

% % --------------- in-the-wild test -------------------%
\begin{wraptable}{r}{0.48\textwidth}
    \centering
    \vspace{-11mm}
    \caption{\textbf{In-the-wild ASD performance on the WASD dataset.} Our method outperforms other models on in-the-wild dataset, demonstrating superior generalization and robustness.}
    \label{tab:wasd}
    \setlength{\tabcolsep}{15pt}
    \renewcommand{\arraystretch}{0.7}
    \scriptsize{
    \begin{tabular}{l|c}
      \toprule
      Model & WASD (mAP)\\
      \midrule
      TalkNet~\cite{talknet}   & 78.4 \\
      ADENet~\cite{adenet}    & 85.6 \\
      Light-ASD~\cite{lightasd} & 85.3 \\
      \hline
      \textbf{Ours} & \textbf{86.1} \\
      \bottomrule
    \end{tabular}
    }
\vspace{-7mm}
\end{wraptable}

% --------------- in-the-wild test (POSTER)-------------------%
% \begin{wraptable}{r}{0.48\textwidth}
%     \centering
%     \vspace{-11mm}
%     \caption{\textbf{In-the-wild ASD performance on the WASD dataset.} Our method outperforms other models on in-the-wild dataset, demonstrating superior generalization and robustness.}
%     \label{tab:wasd}
%     \setlength{\tabcolsep}{15pt}
%     \renewcommand{\arraystretch}{0.7}
%     \scriptsize{
%     \begin{tabular}{l|c}
%       \toprule
%       Model & mAP (\%)\\
%       \midrule
%       TalkNet   & 78.4 \\
%       ADENet   & 85.6 \\
%       Light-ASD& 85.3 \\
%       \textbf{Ours} & \textbf{86.1} \\
%       \bottomrule
%     \end{tabular}
%     }
% \vspace{-7mm}
% \end{wraptable}

\vspace{-5mm}
\subsection{In-the-Wild Results on WASD}
\vspace{-2mm}
To evaluate cross-dataset generalization, we directly applied models trained on AVA-ActiveSpeaker to the WASD dataset without any fine-tuning. As shown in \cref{tab:wasd}, our method achieves 86.1\% mAP, outperforming the strongest baseline ADENet~\cite{adenet} by $+0.5\%$ and Light-ASD~\cite{lightasd} by $+0.8\%$.
In contrast, our lightweight model generalizes more effectively to in-the-wild conditions, suggesting that the proposed consistency regularization encourages the learning of more transferable and modality-robust representations rather than dataset-specific shortcuts. These results demonstrate that multi-level consistency constraints enhance not only robustness to corruption but also cross-domain generalization, even when both modalities remain clean while the scene distribution shifts significantly from the training data.

% --------------- audio only corruption (MUSAN) -------------------%
\begin{table}[t]
\caption{\textbf{Noise Robustness Evaluation on MUSAN.} Comparison of performance (mAP) across three noise types (Babble, Music, Natural) at SNR levels of \{-10, -5, 0, 5, 10\} dB. AVG denotes the average mAP across all SNR levels per noise type.}
\label{tab:musan}
\vspace{-3mm}
\centering
\resizebox{\textwidth}{!}{
\begin{tabular}{l|cccccc|cccccc|cccccc}
\hline
\multirow{2}{*}{Method}
& \multicolumn{6}{c|}{Babble, SNR (dB)} 
& \multicolumn{6}{c|}{Music, SNR (dB)} 
& \multicolumn{6}{c}{Natural, SNR (dB)} \\

& -10 & -5 & 0 & 5 & 10 & AVG
& -10 & -5 & 0 & 5 & 10 & AVG
& -10 & -5 & 0 & 5 & 10 & AVG \\
\hline

TalkNet~\cite{talknet}  
& 85.5 & 87.6 & 89.3 & 90.5 & 91.2 & 88.8
& 84.1 & 86.5 & 88.5 & 90.0 & 91.0 & 88.0
& 85.2 & 87.4 & 89.1 & 90.2 & 90.9 & 88.6\\

ADENet~\cite{adenet}
& 84.0 & 85.8 & 86.7 & 87.5 & 88.4 & 86.5
& 82.5 & 84.5 & 85.9 & 87.3 & 88.4 & 85.7
& 83.6 & 85.3 & 86.4 & 87.4 & 88.3 & 86.2\\

Light-ASD~\cite{lightasd}
& 88.0 & 89.7 & 91.1 & 92.1 & 92.7 & 90.7
& 86.8 & 88.7 & 90.4 & 91.6 & 92.5 & 90.0
& 87.6 & 89.4 & 90.8 & 91.8 & 92.5 & 90.4\\

\hline
\textbf{Ours}
& \textbf{88.3} & \textbf{89.9} & \textbf{91.3} & \textbf{92.3} & \textbf{93.0} & \textbf{91.0}
& \textbf{87.5} & \textbf{89.2} & \textbf{90.8} & \textbf{92.0} & \textbf{92.8} & \textbf{90.5}
& \textbf{88.3} & \textbf{89.9} & \textbf{91.2} & \textbf{92.2} & \textbf{92.8} & \textbf{90.9} \\
\hline

\end{tabular}
}
\vspace{-3mm}
\end{table}

% --------------- audio only corruption (DEMAND) -------------------%
\begin{table}[t]
\caption{\textbf{Real-World Noise Robustness Evaluation on DEMAND.} Performance (mAP) across eight real-world environments spanning outdoor (PARK, RIVER), indoor (CAFE, RESTAURANT, CAFETERIA, MEETING), and transit (METRO, STATION) conditions. AVG is the mean mAP.}
\label{tab:demand}
\centering
\vspace{-3mm}
\scriptsize
\begin{tabular}{l|c|c|c|c|c|c|c|c|c}
\toprule
Method 
& PARK 
& RIVER 
& CAFE 
& RESTAU 
& CAFETER 
& METRO 
& STATION 
& MEETING 
& AVG \\
\midrule
TalkNet~\cite{talknet}   & 90.3 & 89.1 & 86.8 & 86.5 & 86.8 & 91.1 & 89.6 & 85.7 & 88.3 \\
ADENet~\cite{adenet}    & 88.9 & 87.1 & 85.3 & 84.9 & 85.3 & 89.1 & 87.5 & 87.5 & 86.5 \\
Light-ASD~\cite{lightasd} & 92.1 & 91.0 & 89.0 & 88.8 & 89.0 & 92.6 & 91.2 & 87.6 & 90.2 \\
\hline
\textbf{Ours}  & \textbf{92.5} & \textbf{91.3} & \textbf{89.7} & \textbf{89.4} & \textbf{89.7} & \textbf{92.7} & \textbf{91.5} & \textbf{88.9} & \textbf{90.7} \\
\bottomrule
\end{tabular}
\vspace{-6mm}
\end{table}

\subsection{Corruption Robustness Evaluation}
\label{sec:corruption}
% TODO : corruption type 이미지 설명
We evaluated robustness under audio, visual, and joint audio-visual corruptions applied exclusively at test time. All results were reported on the AVA-ActiveSpeaker validation set, and no corrupted data was used during training.

\noindent\textbf{Audio Corruptions.}
\cref{tab:musan,tab:demand} present the results under MUSAN and DEMAND noise conditions, respectively. Our method consistently outperforms the strongest baseline, achieving average mAP gains up to +0.5\% on both MUSAN and DEMAND compared to Light-ASD~\cite{lightasd}. Although the proposed consistency regularization primarily targets cross-modal representation structure rather than audio-specific robustness, the improved inter-modality alignment also contributes to more stable representations under acoustic degradation.

% --------------- visual only corruption -------------------%
\begin{wraptable}{r}{0.5\textwidth}
    \vspace{-11mm}
\caption{\textbf{Visual Corruption Robustness Evaluation.} Performance (mAP) under two visual corruption types: object occlusion with background noise (Object Occ. + Noise) and pixelated face degradation (Pixelated Face).}
\label{tab:visual}
\centering
  \setlength{\tabcolsep}{0pt}
  \renewcommand{\arraystretch}{0.7}
  \scriptsize
\begin{tabular}{l|c|c}
\toprule
Method & Object Occ. + Noise  & Pixelated \\
\midrule
TalkNet~\cite{talknet}   & 70.73 & 92.12 \\
ADENet~\cite{adenet}   & 66.36 &  89.04  \\
Light-ASD~\cite{lightasd} & 76.86 & 93.15 \\
\hline
\textbf{Ours}      & \textbf{78.90} & \textbf{93.47} \\
\bottomrule
\end{tabular}
    \vspace{-7mm}
\end{wraptable}

\noindent\textbf{Visual Corruptions.}
\cref{tab:visual} presents the results under visual-only corruption conditions. Our method achieves +2.04\% mAP improvement under object occlusion, confirming that inter-modality alignment enables the audio stream to compensate for degraded visual input while intra-modality consistency enhances representation robustness. Under pixelation, our method also shows a marginal gain of +0.32\% mAP, maintaining competitive performance even under global low-frequency visual degradations.

% --------------- audio-visual corruption (MUSAN) -------------------%
\begin{table}[t]
\caption{\textbf{Audio-Visual Joint Corruption Robustness Evaluation on MUSAN.} Performance (mAP) under simultaneous audio and visual corruptions, combining MUSAN noise types (Babble, Music, Natural) at SNR levels of \{-10, -5, 0, 5, 10\} dB with two visual degradations: object occlusion + noise (Object Occ.) and face pixelation (Pixelated).}
\label{tab:joint_musan}
\centering
\vspace{-3mm}
\small
\resizebox{\textwidth}{!}{
\begin{tabular}{c|l|cccccc|cccccc|cccccc}
\hline
\multirow{2}{*}{}
& \multirow{2}{*}{Method}
& \multicolumn{6}{c|}{Babble, SNR (dB)}
& \multicolumn{6}{c|}{Music, SNR (dB)}
& \multicolumn{6}{c}{Natural, SNR (dB)}\\

&
& -10 & -5 & 0 & 5 & 10 & AVG
& -10 & -5 & 0 & 5 & 10 & AVG
& -10 & -5 & 0 & 5 & 10 & AVG\\
\hline

% ---------------- (a) ----------------
\multirow{5}{*}[5pt]{\rotatebox{90}{\tiny{Object Occ.}}}
& TalkNet~\cite{talknet}   & 53.7 & 58.8 & 63.1 & 66.3 & 68.1 & 62.0
& 53.8 & 59.3 & 64.0 & 67.2 & 69.1 & 62.7
& 54.1 & 59.3 & 63.7 & 66.8 & 68.9 & 62.6\\

& ADENet~\cite{adenet}   & 55.8 & 59.1 & 61.0 & 62.7 & 64.7 & 60.7
& 52.8 & 56.4 & 59.1 & 62.2 & 64.6 & 59.0
& 54.6 & 57.5 & 59.7 & 62.0 & 64.2 & 59.6\\

& Light-ASD~\cite{lightasd}  & 65.7 & 69.4 & 72.7 & 75.1 & 76.3 & 71.9
& 64.8 & 68.9 & 72.6 & 75.0 & 76.3 & 71.5
& 65.1 & 68.8 & 72.2 & 74.4 & 75.8 & 71.3\\

\cline{2-20}
& \textbf{Ours} & \textbf{66.6} & \textbf{70.2} & \textbf{73.7} & \textbf{76.3} & \textbf{77.8} & \textbf{72.9}
& \textbf{65.8} & \textbf{69.6} & \textbf{73.4} & \textbf{76.1} & \textbf{77.6} & \textbf{72.5}
& \textbf{66.6} & \textbf{70.4} & \textbf{73.8} & \textbf{76.1} & \textbf{77.6} & \textbf{73.0}\\

\hline

% ---------------- (b) ----------------
\multirow{5}{*}[5pt]{\rotatebox{90}{\tiny{Pixelated}}}
& TalkNet~\cite{talknet}  & 85.0 & 87.4 & 89.1 & 90.3 & 91.1 & 88.6
& 83.9 & 86.2 & 88.2 & 89.8 & 90.8 & 87.8
& 84.9 & 87.2 & 88.9 & 90.1 & 90.9 & 88.4\\

& ADENet~\cite{adenet}   & 84.1 & 85.9 & 86.8 & 87.3 & 88.2 & 86.5
& 82.2 & 84.2 & 85.7 & 87.2 & 88.3 & 85.5
& 83.2 & 85.0 & 86.2 & 87.3 & 88.2 & 86.0\\

& Light-ASD~\cite{lightasd} & 87.5 & 89.2 & 90.5 & 91.5 & 92.1 & 90.2
& 86.0 & 87.9 & 89.6 & 90.9 & 91.8 & 89.3
& 86.9 & 88.7 & 90.1 & 91.2 & 91.9 & 89.8\\

\cline{2-20}
& \textbf{Ours} & \textbf{87.9} & \textbf{89.6} & \textbf{90.9} & \textbf{91.9} & \textbf{92.5} & \textbf{90.6}
& \textbf{86.7} & \textbf{88.6} & \textbf{90.2} & \textbf{91.5} & \textbf{92.4} & \textbf{89.9}
& \textbf{87.7} & \textbf{89.4} & \textbf{90.8} & \textbf{91.8} & \textbf{92.4} & \textbf{90.4}\\

\hline
\end{tabular}
}
\vspace{-3mm}
\end{table}

% --------------- audio-visual corruption (DEMAND) -------------------%
\begin{table}[t]
\centering
\caption{\textbf{Audio-Visual Joint Corruption Robustness Evaluation on DEMAND.} Performance (mAP) under simultaneous audio and visual corruptions, combining eight real-world acoustic environments from the DEMAND dataset with two visual degradations: object occlusion with background noise (Object Occlusion + Noise) and face pixelation (Pixelated Face).}
\label{tab:joint_demand}
\vspace{-3mm}
\resizebox{\textwidth}{!}{
\begin{tabular}{l|ccccccccc|ccccccccc}
\hline
& \multicolumn{9}{c|}{Object Occlusion + Noise}
& \multicolumn{9}{c}{Pixelated Face} \\

Method
& \rotatebox{60}{PARK}
& \rotatebox{60}{RIVER}
& \rotatebox{60}{CAFE}
& \rotatebox{60}{RESTO}
& \rotatebox{60}{CAFETER}
& \rotatebox{60}{METRO}
& \rotatebox{60}{STATION}
& \rotatebox{60}{MEETING}
& \rotatebox{60}{AVG}

& \rotatebox{60}{PARK}
& \rotatebox{60}{RIVER}
& \rotatebox{60}{CAFE}
& \rotatebox{60}{RESTO}
& \rotatebox{60}{CAFETER}
& \rotatebox{60}{METRO}
& \rotatebox{60}{STATION}
& \rotatebox{60}{MEETING}
& \rotatebox{60}{AVG} \\
\hline

TalkNet~\cite{talknet}
& 67.5 & 63.0 & 60.4 & 57.6 & 60.4 & 69.0 & 63.9 & 59.1 & 62.6
& 90.0 & 89.0 & 86.6 & 86.1 & 86.6 & 90.8 & 89.4 & 85.4 & 88.0 \\

ADENet~\cite{adenet}
& 66.2 & 61.8 & 61.1 & 59.0 & 61.1 & 66.0 & 62.3 & 60.0 & 62.2
& 88.8 & 87.2 & 85.0 & 84.5 & 85.0 & 89.0 & 87.5 & 83.6 & 86.3 \\

Light-ASD~\cite{lightasd}
& 76.0 & 69.7 & 71.7 & 68.8 & 71.7 & 75.4 & 70.3 & 68.4 & 71.5
& 91.5 & 90.3 & 88.4 & 88.1 & 88.4 & 92.0 & 90.5 & 86.7 & 89.5 \\

\hline
\textbf{Ours}
& \textbf{76.1} & \textbf{71.2} & \textbf{71.8} & \textbf{70.0} & \textbf{71.8} & \textbf{76.6} & \textbf{72.4} & \textbf{70.1} & \textbf{72.5}
& \textbf{92.0} & \textbf{90.8} & \textbf{89.1} & \textbf{88.9} & \textbf{89.1} & \textbf{92.3} & \textbf{91.0} & \textbf{88.3} & \textbf{90.2} \\

\hline
\end{tabular}
}
\vspace{-5mm}
\end{table}

\noindent\textbf{Audio-Visual Joint Corruptions.}
\cref{tab:joint_musan} and~\cref{tab:joint_demand} present results under simultaneous audio-visual corruptions, representing the most challenging and realistic degradation scenario.
Under joint corruption with object occlusion, the proposed method yields consistent improvements: $+1.23\%$ average mAP with MUSAN noise and $+1.0\%$ with DEMAND noise.
The improvement is most pronounced at severe SNR levels ($-10$ dB), suggesting that consistency regularization is especially effective when both modalities are heavily degraded simultaneously. This indicates that the multi-level consistency losses create complementary redundancy between modalities, allowing partial information from each corrupted stream to jointly support correct predictions.
For pixelation-based joint corruptions, our method also maintains a consistent edge over the baseline ($+0.53\%$ on MUSAN, $+0.7\%$ on DEMAND), consistent with the visual-only pixelation results.

  \vspace{-3mm}
\subsection{Analysis of Consistency Losses}
  \vspace{-2mm}
  
\begin{wrapfigure}{r}{0.48\textwidth}
  \vspace{-11mm}
  \centering
  \includegraphics[width=1.0\linewidth]{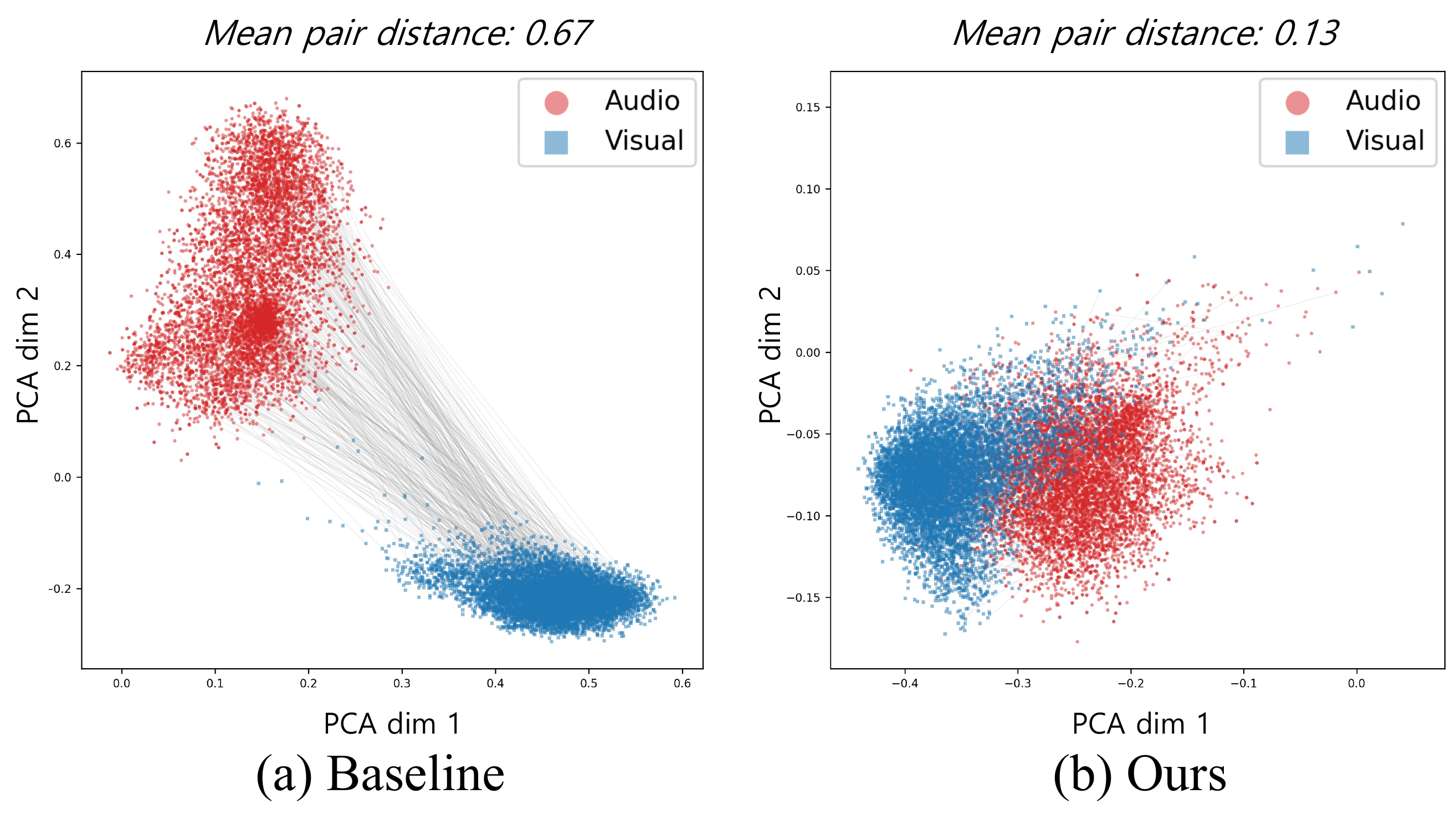}
  \vspace{-7mm}
  \caption{\textbf{PCA visualization of audio and visual embeddings.} Decreased mean paired distance indicates improved cross-modal alignment and intra-modal compactness.}
  \label{fig:pca_alignment}
  \vspace{-8mm}
\end{wrapfigure}

\begin{figure*}[t]
    \begin{minipage}[t]{\linewidth}
	\centering
  \includegraphics[width=1.0\linewidth]{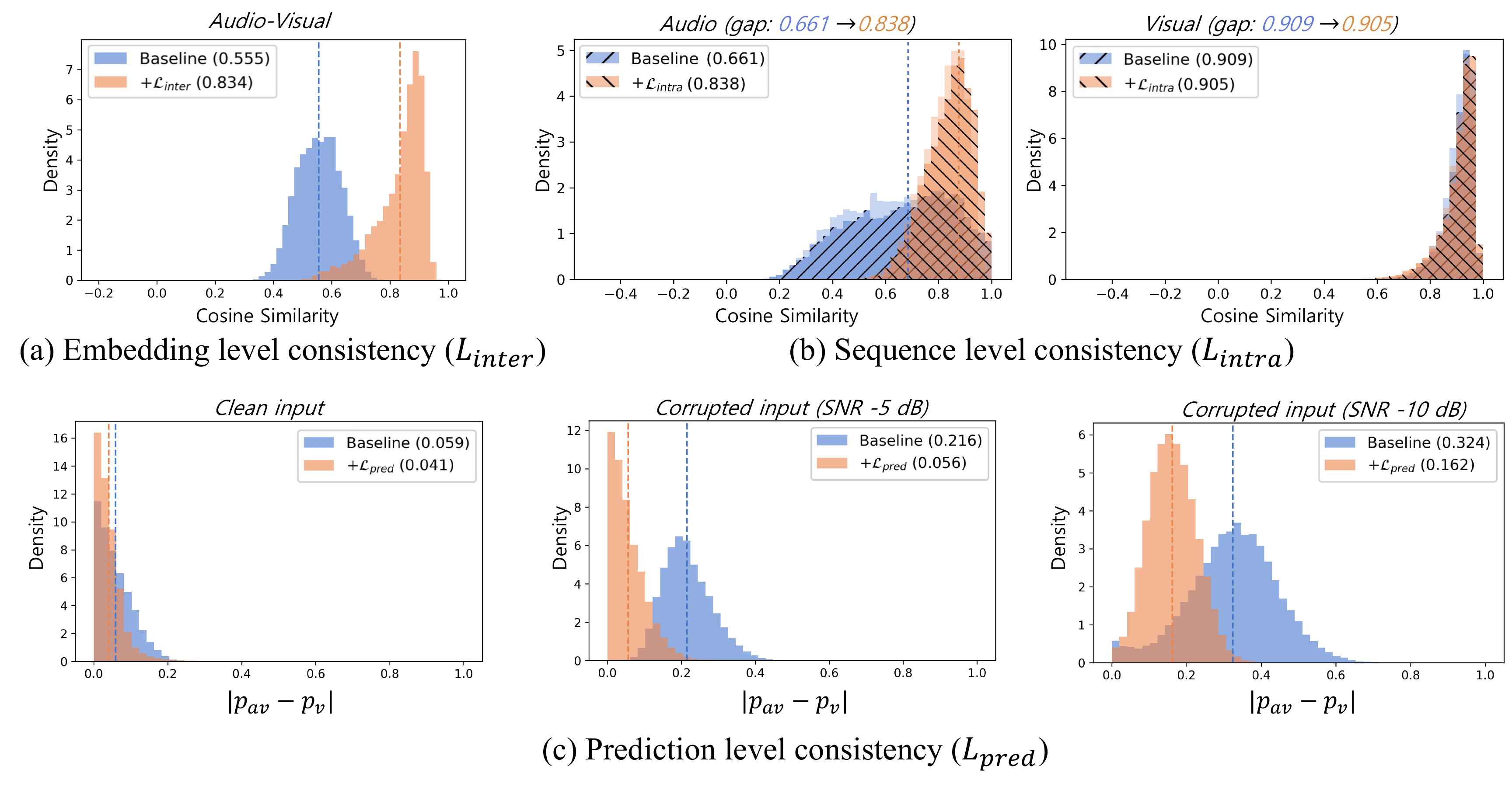}
    \end{minipage}
  \vspace{-7mm}
  \caption{
  \textbf{Analysis of each consistency terms.}(a) $\mathcal{L}_{\text{inter}}$ substantially increases cosine similarity between audio and visual embeddings. (b) $\mathcal{L}_{\text{intra}}$ tightens within-track clustering, with a more pronounced effect on the audio stream. (c) $\mathcal{L}_{\text{pred}}$ maintains consistent AV/V prediction agreement even under corruption.}
  \label{fig:consistency}
    \vspace{-5mm}
\end{figure*}

\cref{fig:pca_alignment} visualizes audio and visual embeddings projected onto a shared 2D PCA space computed on the AVA validation set. In the baseline (\cref{fig:pca_alignment}(a)), audio and visual embeddings occupy entirely disjoint regions of the feature space, with loosely scattered within-modality distributions. In contrast, our model (\cref{fig:pca_alignment}(b)) exhibits markedly different behavior: the two modalities are co-located within a compact shared region, with the mean paired distance dropping from $0.67$ to $0.13$ ($-80.6\%$). Moreover, embeddings within each modality form tighter clusters, indicating that consistency regularization simultaneously promotes cross-modal alignment and intra-modal compactness. We examine the individual contribution of each consistency loss, as follows:

\noindent\textbf{Inter-Modality Consistency ($\mathcal{L}_{\text{inter}}$).}
\cref{fig:consistency}(a) shows the distribution of cosine similarity between audio and visual embeddings. Without $\mathcal{L}_{\text{inter}}$, the baseline produces a mean cosine similarity of 0.555, indicating only moderate cross-modal alignment despite element-wise fusion. Introducing $\mathcal{L}_{\text{inter}}$ shifts the distribution substantially to the right, increasing the mean to 0.834 (+0.279) and concentrating most of the mass above 0.8. This result confirms that $\mathcal{L}_{\text{inter}}$ is the primary driver of the geometric alignment observed in \cref{fig:pca_alignment}(b).

\noindent\textbf{Intra-Modality Consistency ($\mathcal{L}_{\text{intra}}$).}
\cref{fig:consistency}(b) reports pairwise cosine similarity between same-track frame pairs within each modality. For the audio stream, $\mathcal{L}_{\text{intra}}$ improves the inter-class similarity gap from 0.661 to 0.838 and reduces intra-track embedding variance from 0.039 to 0.017, indicating tighter within-speaker clustering. The effect on the visual stream is smaller (0.005$\to$0.003), which is expected since visual appearance is already highly consistent within a track due to the face-crop pipeline. This asymmetry suggests that $\mathcal{L}_{\text{intra}}$ primarily benefits the noisier audio modality and accounts for the improved per-modality compactness observed in \cref{fig:pca_alignment}(b).

\noindent\textbf{Prediction-Level Consistency ($\mathcal{L}_{\text{pred}}$).}
\cref{fig:consistency}(c) compares the AV/V-only prediction disagreement $|p_{av} - p_v|$ on clean and corrupted inputs (Gaussian noise, SNR $-5$ dB and $-10$ dB). On clean data, both models show similarly low disagreement (Baseline: 0.059, $+\mathcal{L}_{\text{pred}}$: 0.041), indicating that the two streams already produce largely consistent predictions. However, in corruption, the baseline disagreement increases sharply to 0.216 and 0.324 at SNR $-5$ dB and $-10$ dB, respectively, as noisy audio conflicts with visual predictions. In contrast, with $\mathcal{L}_{\text{pred}}$ the distribution collapses toward zero (0.056 and 0.162), indicating that the model suppresses unreliable audio cues and maintains consistent predictions under degraded inputs. This selective behavior explains why $\mathcal{L}_{\text{pred}}$ improves robustness without sacrificing clean mAP: the loss provides negligible additional signal when predictions are already consistent, but actively suppresses modality conflicts under corruption.

\subsection{Ablation Study}
To analyze the contribution of each consistency loss, we conducted an ablation study on the AVA-ActiveSpeaker validation set.
\cref{tab:ablation} reports the mAP of the baseline and each variant obtained by selectively enabling the inter-modality, intra-modality, and prediction-level consistency losses.

% --------------- ablation test -------------------%
\begin{wraptable}{r}{0.5\textwidth} 
\centering
\vspace{-11mm}
\caption{\textbf{Ablation study on different training components.} 
We evaluate the contribution of \textit{inter}, \textit{intra}, and \textit{pred} objectives on AVA (mAP (\%)).}
\label{tab:ablation}
\setlength{\tabcolsep}{5pt}
\renewcommand{\arraystretch}{0.7}
\label{sec:experiments}
\scriptsize
\begin{tabular}{ccc|c}
\toprule
Inter & Intra & Pred & mAP \\
\midrule
\xmark & \xmark & \xmark & 93.61 \\

\cmark & \xmark & \xmark & 93.70 \\
\xmark & \cmark & \xmark & 93.62 \\
\xmark & \xmark & \cmark & 93.68 \\

\cmark & \cmark & \cmark & \textbf{93.80} \\
\bottomrule
\end{tabular}
    \vspace{-7mm}
\end{wraptable}

Each individual consistency loss provides a complementary improvement over the baseline (93.61\%).
Inter-modality consistency yields the largest individual gain ($+0.09\%$), confirming that explicit cross-modal alignment is the most direct mechanism for improving robustness.
Intra-modality consistency provides a modest gain ($+0.01\%$), as within-modality cluster separation alone is insufficient without cross-modal anchoring, and prediction-level consistency achieves $+0.07\%$ by stabilizing unimodal predictions via knowledge distillation.
When all three losses are combined, the model achieves 93.80\% mAP, demonstrating that the three constraints are complementary and mutually reinforcing. The modest clean-data gains are expected, as the primary benefit of consistency regularization manifests under corrupted conditions, as evidenced by the larger improvements in \cref{sec:corruption}.

%\subsection{Generalization to Other Backbones}
%To verify that the proposed consistency losses generalize beyond Light-ASD, we applied $C^3$ to TalkNet and ADENet without any architecture-specific modification. As shown in Tab.\ref{tab:generalization}, consistent improvements are observed across all corruption conditions for both backbones, demonstrating that the proposed losses are architecture-agnostic and can be seamlessly integrated into existing ASD frameworks.
%\begin{table}[h]
%\vspace{-3mm}
%\centering
%\caption{\textbf{Generalization across backbones (mAP \%)}.}
%\label{tab:generalization}
%\resizebox{\linewidth}{!}{%
%\begin{tabular}{lcccccccc}
%\toprule
%\multirow{2}{*}{Method} & Audio & Audio & Visual & MUSAN + & MUSAN + & DEMAND + & DEMAND + \\
% & (MUSAN) & (DEMAND) & (Basic) & Obj. Occ. & Pix. & Obj. Occ. & Pix. \\
%\midrule
%TalkNet      & 88.47 & 88.25 & 81.52 & 62.41 & 88.25 & 62.62 & 88.00 \\
%TalkNet+$C^3$                       & \textbf{88.66} & \textbf{88.52} & \textbf{84.44} & \textbf{68.29} & \textbf{88.54} & \textbf{68.14 }& \textbf{88.01} \\
%\midrule
%ADENet        & 84.58 & 86.50 & 77.70 & 59.81 & 84.73 & 61.07 & 85.12 \\
%ADENet+$C^3$                        &  \textbf{85.05}  &  \textbf{86.63}  & \textbf{81.31}  & \textbf{61.88} & \textbf{85.98} &   \textbf{61.88} & \textbf{86.41} \\
%\bottomrule
%\end{tabular}}
%\vspace{-3mm}
%\end{table}

\section{Conclusion}
In this work, we proposed $C^3$ASD, a multi-level consistency-driven framework for robust active speaker detection. We identify the lack of explicit consistency constraints as a key limitation of existing ASD models under real-world corruptions and address it with three complementary losses: embedding-level inter-modality consistency, intra-modality consistency, and prediction-level consistency via knowledge distillation. Experiments on AVA-ActiveSpeaker and WASD demonstrate improved robustness under diverse audio, visual, and joint corruptions, with notable gains under object occlusion and severe SNR conditions. The proposed losses are lightweight, require no additional annotations or corrupted training data, and can be easily integrated into existing ASD architectures, encouraging future research on representation-level regularization for robust multimodal learning.

\section*{Acknowledgements}
This work was supported in part by the National Research Foundation of Korea(NRF) grant funded by the Korea government(MSIT) (RS-2025-02217071), and in part by the Institute of Information \& communications Technology Planning \& Evaluation (IITP) grant funded by the Korea government(MSIT) (RS-2021-II211341, RS-2025-25422680).
% % ---- Bibliography ----
% %
% % BibTeX users should specify bibliography style 'splncs04'.
% % References will then be sorted and formatted in the correct style.
% %

\bibliographystyle{splncs04}
\bibliography{main}
\clearpage
\setcounter{page}{1}
 \setcounter{equation}{0}
 \renewcommand{\theequation}{A.\arabic{equation}}
 \setcounter{figure}{0}
 \renewcommand{\thefigure}{A.\arabic{figure}}
 \setcounter{table}{0}
 \renewcommand{\thetable}{A.\arabic{table}}
  \setcounter{section}{0}
\renewcommand{\thesection}{A.\arabic{section}}

% \begin{center}
%     {\Large \textbf{Supplementary Material for\\[2pt]
%     ``$C^3$ASD: Multi-Level Consistency-Driven Representation Learning for Robust Active Speaker Detection''}}
%     \vspace{6mm}
% \end{center}

%\section*{Supplementary Material}

\section{Complementary Effect of $\mathcal{L}_{\text{inter}}$ and $\mathcal{L}_{\text{intra}}$}

\begin{figure}[th]
    \vspace{-3mm}
    \begin{minipage}[t]{\linewidth}
	\centering
  \includegraphics[width=1.0\linewidth]{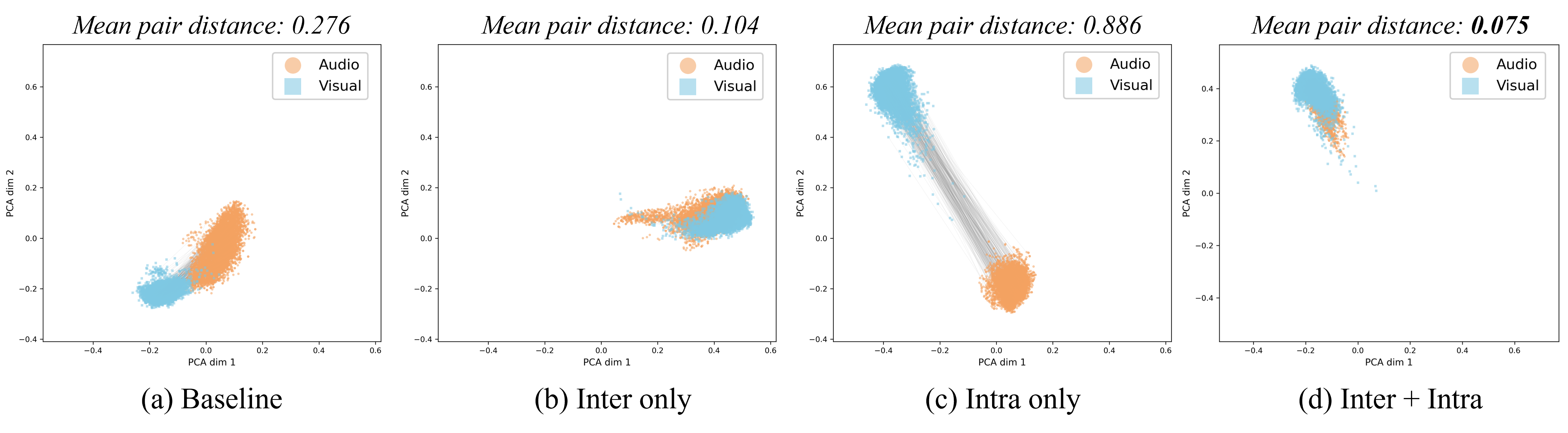}
    \end{minipage}
  \caption{\textbf{PCA visualization of audio and visual embeddings under four training configurations}. Each gray line connects the audio and visual embeddings of the same video segment. Mean paired distance between corresponding audio-visual embeddings is reported above each plot.}
    \label{fig:pca}
    \vspace{-3mm}
\end{figure}

Fig.~\ref{fig:pca} visualizes audio and visual embeddings in a shared 2D PCA space to motivate why both losses are necessary. The baseline (a) shows loosely distributed embeddings with a mean paired distance of 0.276, as each encoder develops its own geometry without any consistency constraint. With $\mathcal{L}_{\text{inter}}$ alone (b), the mean paired distance drops to 0.104, confirming effective cross-modal alignment, but both distributions remain diffuse without tight internal structure; making embeddings near the decision boundary susceptible to corruption-induced shifts.

With $\mathcal{L}_{\text{intra}}$ alone (c), each modality forms a compact internal cluster compared to the baseline, which is its intended effect; however, without any cross-modal signal, the two modalities remain in entirely different regions of the space (mean distance 0.886), so the element-wise fusion $\mathbf{e}_{av} = \mathbf{e}_a + \mathbf{e}_v$ simply adds embeddings from misaligned spaces, and the clean modality cannot steer the fused representation toward the correct decision boundary. 

Combining both losses (d) resolves both failure modes simultaneously: the two modalities are brought into a shared region (mean distance 0.075) while each modality maintains compact internal clustering, so the clean modality can anchor the fused representation and the enlarged cluster margin absorbs corruption-induced drift.

\section{Theoretical Analysis of Consistency Losses}

We provide formal derivations for each consistency loss in $C^3$ASD. we first state the question the derivation addresses and then present the corresponding mathematical argument.

% ─────────────────────────────────────────────────────────────────────────
\subsection{Inter-Modality Consistency Loss ($\mathcal{L}_{\text{inter}}$)}

\paragraph{Questions.}
The main paper claims that aligning audio and visual embeddings via cosine
similarity improves robustness to corruption.
This raises two questions that require justification:
\begin{enumerate}
    \item \emph{Does alignment actually help under corruption?}
          If the audio embedding drifts due to noise,
          why does aligning it with the visual embedding help?
    \item \emph{Does minimizing $\mathcal{L}_{\text{inter}}$ risk collapsing
          all embeddings to a single point?}
          A constant embedding trivially achieves
          $\mathcal{L}_{\text{inter}} = 0$,
          so we must show that this degenerate solution is avoided.
\end{enumerate}
We first derive the gradient of $\mathcal{L}_{\text{inter}}$ to characterize how it modifies the embedding space, then use this to answer both questions.

\noindent\textbf{Gradient derivation.}
Let $\hat{\mathbf{a}}^{(i)} = \mathbf{e}_a^{(i)}/\left\|\mathbf{e}_a^{(i)}\right\|$ and
$\hat{\mathbf{v}}^{(i)} = \mathbf{e}_v^{(i)}/\left\|\mathbf{e}_v^{(i)}\right\|$.
For a single speaking frame $i \in \mathcal{S}$, its contribution to
$\mathcal{L}_{\text{inter}}$ is:
\begin{equation}
    \ell_i = 1 - \frac{\mathbf{e}_a^{(i)} \cdot \mathbf{e}_v^{(i)}}
               {\left\|\mathbf{e}_a^{(i)}\right\|\,\left\|\mathbf{e}_v^{(i)}\right\|}.
\end{equation}
Applying the quotient rule with respect to $\mathbf{e}_a^{(i)}$:
\begin{equation}
    \frac{\partial \ell_i}{\partial \mathbf{e}_a^{(i)}}
    = -\frac{1}{\left\|\mathbf{e}_a^{(i)}\right\|}
      \left[\hat{\mathbf{v}}^{(i)}
      - (\hat{\mathbf{a}}^{(i)} \cdot \hat{\mathbf{v}}^{(i)})\,
      \hat{\mathbf{a}}^{(i)}\right].
    \label{eq:grad_inter}
\end{equation}
The bracketed term is the rejection of $\hat{\mathbf{v}}^{(i)}$ from
$\hat{\mathbf{a}}^{(i)}$, \ie the component of $\hat{\mathbf{v}}^{(i)}$
orthogonal to $\hat{\mathbf{a}}^{(i)}$:
\begin{equation}
    \left[\hat{\mathbf{v}}^{(i)}
    - (\hat{\mathbf{a}}^{(i)}\cdot\hat{\mathbf{v}}^{(i)})
    \hat{\mathbf{a}}^{(i)}\right] \cdot \hat{\mathbf{a}}^{(i)} = 0.
\end{equation}
Since the gradient is orthogonal to $\mathbf{e}_a^{(i)}$, gradient descent \emph{rotates} $\mathbf{e}_a^{(i)}$ toward $\mathbf{e}_v^{(i)}$ while preserving $\left\|\mathbf{e}_a^{(i)}\right\|$ to first order $\left(\left\|\mathbf{e}_a^{(i)} - \eta\mathbf{g}\right\|^2= \left\|\mathbf{e}_a^{(i)}\right\|^2 + O(\eta^2)\right)$. That is, $\mathcal{L}_{\text{inter}}$ aligns directions without distorting the magnitude of each encoder's output.

\noindent\textbf{Answer to~(1): Robustness under corruption.}
Because $\mathcal{L}_{\text{inter}}$ aligns both modalities toward each other, the two embeddings reinforce each other in the fused representation $\mathbf{e}_{av}^{(i)} = \mathbf{e}_a^{(i)} + \mathbf{e}_v^{(i)}$. When audio corruption shifts $\mathbf{e}_a^{(i)}$ to $\tilde{\mathbf{e}}_a^{(i)} = \mathbf{e}_a^{(i)} + \boldsymbol{\delta}$, the angular deviation of the fused vector is bounded by:
\begin{equation}
    \angle\left(\mathbf{e}_{av}^{(i)},\, \tilde{\mathbf{e}}_{av}^{(i)}\right)
    \leq \arctan\frac{\|\boldsymbol{\delta}\|}
    {\left\|\mathbf{e}_a^{(i)} + \mathbf{e}_v^{(i)}\right\|},
\end{equation}
which decreases as $\left\|\mathbf{e}_{av}^{(i)}\right\|$ grows. Since alignment ensures $\mathbf{e}_a^{(i)}$ and $\mathbf{e}_v^{(i)}$ point in similar directions, their sum has large magnitude, geometrically suppressing the effect of $\boldsymbol{\delta}$ on the fused vector's direction.

Crucially, this alignment is applied only to speaking frames ($i \in \mathcal{S}$), where the audio stream carries genuine speech signals correlated with lip movements. Non-speaking frames are excluded because their audio content bears no semantic correspondence to the visual stream, and aligning them would introduce noise into the learned representations.

\noindent\textbf{Answer to~(2): Collapse prevention.} $\mathcal{L}_{\text{inter}}$ alone could drive all embeddings toward a constant $\mathbf{c}$. We show that $\mathcal{L}_{\text{base}}$ prevents this.
Suppose all embeddings collapse to a constant $\mathbf{e}_a^{(i)} = \mathbf{e}_v^{(i)} = \mathbf{c}$ for every frame $i$. Then the linear head produces a fixed logit $\mathbf{z} = W\mathbf{c}+\mathbf{b}$, yielding a constant prediction $p^* = \sigma(\mathbf{z})$ for all frames.
Substituting into the binary cross-entropy gives
\begin{equation}
    \mathcal{L}_{\text{base}}\big|_{\text{collapse}}
    = -\rho\log p^* - (1-\rho)\log(1-p^*),
\end{equation}
where $\rho$ is the empirical speaking rate. This is minimized at $p^*=\rho$, giving the binary entropy $H(\rho) = -\rho\log\rho-(1-\rho)\log(1-\rho) > 0$. Since a sufficiently expressive model can achieve $\mathcal{L}_{\text{base}} \to 0$ by correctly classifying each frame, collapse is strictly suboptimal for any $\lambda_1 > 0$. Thus no negative pairs or stop-gradient tricks are needed; the classification objective alone provides a natural collapse-prevention mechanism.

% ─────────────────────────────────────────────────────────────────────────
\subsection{Intra-Modality Consistency Loss ($\mathcal{L}_{\text{intra}}$)}

\paragraph{Questions.}
$\mathcal{L}_{\text{inter}}$ aligns directions across modalities but does not constrain the geometry within each modality.
Two questions arise:
\begin{enumerate}
    \item \emph{What does minimizing $\mathcal{L}_{\text{intra}}$ optimize geometrically?} The loss is motivated informally as ``separating clusters''; we need a precise characterization.
    \item \emph{Why is the track-aware positive set necessary?}
    Treating all same-label frames as positives is a natural baseline; we need to show why this fails.
\end{enumerate}
We first derive the optimality condition of $\mathcal{L}_{\text{intra}}$ to answer~(1), then trace the effect of naive positive set construction through the gradient to answer~(2).

\noindent\textbf{Answer to~(1): Cluster margin maximization.}
Let $s_{ij} = \bar{\mathbf{e}}_i \cdot \bar{\mathbf{e}}_j / \tau$ and define the softmax probability $p_{ij} = \exp(s_{ij}) / Z(i)$, where $Z(i) = \sum_{k\in\mathcal{N}(i)}\exp(s_{ik})$. Writing $\log p_{ij} = s_{ij} - \log Z(i)$, the per-anchor loss becomes:
\begin{equation}
    \mathcal{L}_i
    = -\frac{1}{|P(i)|}\sum_{j\in P(i)} s_{ij} + \log Z(i).
\end{equation}
Taking the partial derivative with respect to $s_{ij'}$ for
$j' \in P(i)$:
\begin{equation}
    \frac{\partial \mathcal{L}_i}{\partial s_{ij'}}
    = -\frac{1}{|P(i)|} + p_{ij'}.
\end{equation}
Setting this to zero gives $p_{ij'} = 1/|P(i)|$, i.e., all positive pairs achieve equal softmax weight. For a negative sample $k \in \mathcal{N}(i)\setminus P(i)$:
\begin{equation}
    \frac{\partial \mathcal{L}_i}{\partial s_{ik}}
    = p_{ik}.
\end{equation}
Setting this to zero requires $p_{ik} \to 0$, \ie $s_{ik} \to -\infty$, which corresponds to maximal angular separation on $\mathcal{S}^{d-1}$.
Combining both conditions:
\begin{align}
    \frac{\partial\mathcal{L}_{\text{intra}}}
         {\partial s_{ij}} = 0
    \;&\Rightarrow\;
    p_{ij} = \frac{1}{|P(i)|}
    \quad (j\in P(i)), \label{eq:opt_pos} \\
    \frac{\partial\mathcal{L}_{\text{intra}}}
         {\partial s_{ik}} = 0
    \;&\Rightarrow\;
    p_{ik} \to 0
    \quad (k\in\mathcal{N}(i)\setminus P(i)). \label{eq:opt_neg}
\end{align}
Condition~\eqref{eq:opt_pos} requires all positive pairs to form a tight cluster with equal pairwise similarity. Condition~\eqref{eq:opt_neg} pushes negative pairs to maximal angular separation. Minimizing $\mathcal{L}_{\text{intra}}$ is therefore equivalent to maximizing the margin between speaking and non-speaking clusters on $\mathcal{S}^{d-1}$~\cite{wang2020understanding}.

\noindent\textbf{Gradient analysis and connection to Fisher discriminant.} The optimality conditions above characterize the \emph{solution}, but not how gradient descent gets there.
We now derive the gradient with respect to $\bar{\mathbf{e}}_i$. Using the chain rule through the inner products $s_{ik} = \bar{\mathbf{e}}_i \cdot \bar{\mathbf{e}}_k / \tau$:
\begin{align}
    \frac{\partial \mathcal{L}_i}{\partial \bar{\mathbf{e}}_i}
    &= \frac{\partial \log Z(i)}{\partial \bar{\mathbf{e}}_i}
     - \frac{1}{|P(i)|}\sum_{j\in P(i)}
       \frac{\partial s_{ij}}{\partial \bar{\mathbf{e}}_i} \nonumber \\
       &= \frac{1}{\tau}\sum_{k\in\mathcal{N}(i)} p_{ik}\,\bar{\mathbf{e}}_k
     - \frac{1}{\tau}\,\boldsymbol{\mu}^+_i,
\end{align}
giving:
\begin{equation}
    \frac{\partial\mathcal{L}_{\text{intra}}}{\partial\bar{\mathbf{e}}_i}
    = \frac{1}{\tau}\!\left[
        \sum_{k\in\mathcal{N}(i)} p_{ik}\,\bar{\mathbf{e}}_k
        - \boldsymbol{\mu}^+_i
      \right],
    \label{eq:intra_grad}
\end{equation}
where $\boldsymbol{\mu}^+_i = \frac{1}{|P(i)|}\sum_{j\in P(i)}\bar{\mathbf{e}}_j$ is the mean of the positive set. In the limit $\tau\to 0$, $p_{ik}\to 0$ for all $k\notin P(i)$ and $p_{ij}\to 1/|P(i)|$ for $j\in P(i)$, so the positive terms in the sum approach $\boldsymbol{\mu}^+_i$ and cancel, leaving only the negative repulsion terms. The gradient update therefore moves $\bar{\mathbf{e}}_i$ toward $\boldsymbol{\mu}^+_i$ (reducing within-class scatter $S_W$) and away from opposite-class embeddings (increasing between-class separation $\|\boldsymbol{\mu}_1 - \boldsymbol{\mu}_0\|^2$).

This is precisely the gradient ascent direction of the Fisher discriminant ratio~\cite{fisher} $J_F = \|\boldsymbol{\mu}_1 - \boldsymbol{\mu}_0\|^2 / S_W$. The connection is exact for $\ell_2$-normalized embeddings, where cosine distance and squared Euclidean distance are related by:
\begin{equation}
    \|\bar{\mathbf{e}}_i - \bar{\mathbf{e}}_j\|^2
    = 2(1 - \bar{\mathbf{e}}_i \cdot \bar{\mathbf{e}}_j),
\end{equation}
so minimizing cosine distance within a class and maximizing it across classes is equivalent to minimizing $S_W$ and maximizing
$\|\boldsymbol{\mu}_1 - \boldsymbol{\mu}_0\|^2$ in Euclidean space.

\noindent\textbf{Answer to~(2): Necessity of the track-aware constraint.} Consider a batch with speakers $A$ and $B$, both speaking ($y=1$). With the naive positive set $P_{\text{naive}}(i) = \{j\neq i \mid y_j = y_i\}$, the positive mean for any anchor becomes:
\begin{equation}
    \boldsymbol{\mu}^+_{\text{naive}}
    = \frac{1}{2}\bigl(\boldsymbol{\mu}^A_{\text{speak}}
                      + \boldsymbol{\mu}^B_{\text{speak}}\bigr).
\end{equation}
Substituting into~\eqref{eq:intra_grad}, the gradient pulls $\bar{\mathbf{e}}_i$ toward a mixture of two different speakers, a semantically incorrect target that conflates distinct identities. The track-aware constraint $P(i) = \{j\neq i \mid y_j=y_i,\, g_j=g_i\}$ restricts the positive mean to the same speaker:
\begin{equation}
    \boldsymbol{\mu}^+_i \ = \boldsymbol{\mu}^{g_i}_{\text{speak}},
\end{equation}
eliminating cross-speaker interference and ensuring each speaker's representations are structured independently.

% ─────────────────────────────────────────────────────────────────────────
\subsection{Prediction-Level Consistency Loss ($\mathcal{L}_{\text{pred}}$)}

\paragraph{Questions.}
Even with well-structured embedding spaces, unimodal classification heads may remain miscalibrated relative to the multimodal branch.
Two design choices require justification:
\begin{enumerate}
    \item \emph{Why MSE rather than KL divergence},
          which is the standard choice in knowledge distillation~\cite{distilling}?
    \item \emph{Is the confidence mask principled, or heuristic?}
          Filtering by teacher confidence is intuitive, but we need to show it strictly reduces the distillation risk.
\end{enumerate}
We address~(1) by comparing gradient magnitudes in the high-confidence regime selected by the mask, and~(2) by decomposing the expected distillation loss under a teacher noise model.

\noindent\textbf{Answer to~(1): MSE vs. KL divergence.}
For binary predictions $p_m, p_{av}\in[0,1]$, let $\varepsilon = p_m - p_{av}$. Taylor-expanding the KL divergence around $\varepsilon = 0$:
\begin{equation}
    \mathrm{KL}[p_m\,\|\,p_{av}]
    \approx \frac{\varepsilon^2}{2\,p_{av}(1-p_{av})}.
\end{equation}
Comparing gradient magnitudes with respect to $p_m$:
\begin{equation}
    \left|\frac{\partial\,\mathrm{KL}}{\partial p_m}\right|
    \;\xrightarrow{p_{av}(1-p_{av})\,\to\, 0}\; +\infty,
    \qquad
    \left|\frac{\partial\,\mathrm{MSE}}{\partial p_m}\right|
    = 2|\varepsilon| \leq 2.
\end{equation}
The confidence mask selects frames satisfying
$\max(p_{av},\, 1{-}p_{av}) \geq \theta$,
i.e.\ frames where $p_{av}$ is near $0$ or $1$, precisely the regime in which $p_{av}(1{-}p_{av})\to 0$ and KL gradients diverge. MSE remains bounded throughout, making it the appropriate distillation loss under confidence masking, answering~(1).

\noindent\textbf{Answer to~(2): Confidence mask.}
Model the teacher as $p_{av} = p^* + \xi$, where $p^*$ is the true conditional probability and $\xi\sim\mathcal{N}(0,\sigma_t^2)$ is teacher noise. The expected distillation loss decomposes as:
\begin{equation}
    \mathbb{E}_\xi\bigl[(p_m - p_{av})^2\bigr]
    = \underbrace{(p_m - p^*)^2}_{\text{student bias}^2}
    + \underbrace{\sigma_t^2}_{\text{teacher noise}}.
    \label{eq:distill_risk}
\end{equation}
Under the reasonable assumption that high-confidence frames ($\max(p_{av}, 1-p_{av}) \geq \theta$) correspond to lower teacher noise, i.e., $\sigma_t^2(\theta) = \mathrm{Var}[\xi \mid p_{av} \geq \theta] < \sigma_t^2$, applying the law of total variance gives:
\begin{equation}
    \sigma_t^2
    = P(\mathcal{M})\cdot\sigma_t^2(\theta)
    + P(\mathcal{M}^c)\cdot\sigma_t^2(\mathcal{M}^c),
\end{equation}
which is consistent since $\sigma_t^2(\theta) < \sigma_t^2 < \sigma_t^2(\mathcal{M}^c)$. Substituting into~\eqref{eq:distill_risk}:
\begin{equation}
    \mathbb{E}\bigl[(p_m-p_{av})^2 \mid \mathcal{M}(\theta)\bigr]
    = (p_m-p^*)^2 + \sigma_t^2(\theta)
    < (p_m-p^*)^2 + \sigma_t^2.
\end{equation}
The mask strictly reduces the expected distillation risk by filtering frames where the teacher is most uncertain, answering~(2). As training progresses and $\sigma_t^2$ decreases, $|\mathcal{M}|$ grows naturally, providing an implicit curriculum~\cite{curriculum} from easy to hard frames without any explicit scheduling.

\section{Qualitative Results of Active Speaker Detection}

We provide qualitative results on WASD~\cite{wasd} and AVA-ActiveSpeaker~\cite{ava}. In all visualizations, a \textcolor{green}{green} bounding box indicates a predicted active speaker, a \textcolor{red}{red} bounding box indicates a predicted non-speaker, and a \textbf{\textcolor{suppGreen}{green}} border with a checkmark denotes a correctly classified frame. For better visualization, we refer the reader to the demonstration video samples provided in the ``demo videos'' folder of the supplementary material.

\begin{figure*}[h!]
    \begin{minipage}[t]{\linewidth}
	\centering
  \includegraphics[width=1.0\linewidth]{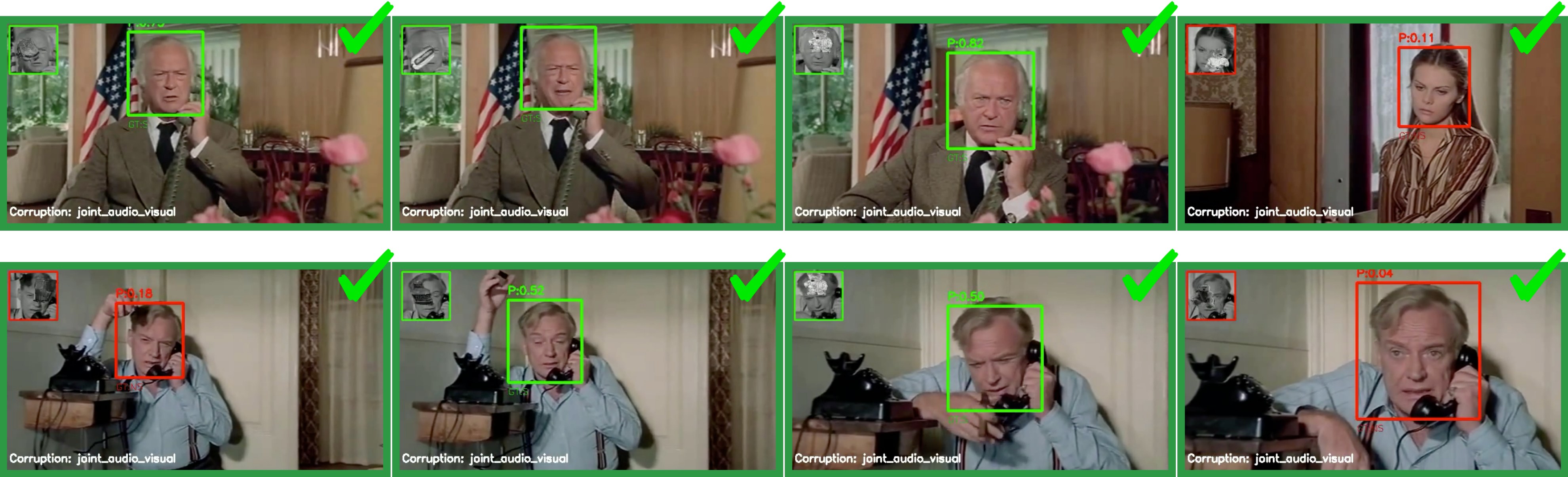}
    \end{minipage}
  \caption{\textbf{Qualitative Results on AVA-ActiveSpeaker under Joint Audio-Visual corruption}. We correctly identifies active speakers across challenging scenes despite simultaneous degradation of both modalities. The COCO patch inserted for visual corruption is visible in the top-right corner of each frame(best viewed zoomed in).}
    \label{fig:supp_ava}
    \vspace{-3mm}
\end{figure*}

Fig.~\ref{fig:supp_wasd} shows results on the WASD dataset without any additional corruption. WASD is inherently a challenging wild dataset collected from diverse real-world scenarios, so we evaluate our model directly on the original data without further degradation. Fig.~\ref{fig:supp_ava} shows results on AVA-ActiveSpeaker under joint audio-visual corruption. Visual corruption is applied by inserting random COCO patches~\cite{coco} at scale 0.4, visible in the top-right region of each frame. Audio corruption is applied by adding Babble noise from the MUSAN~\cite{musan}; please refer to the accompanying demo videos to observe the audio corruption. Despite simultaneous degradation of both modalities, $C^3$ASD correctly identifies active speakers in all displayed frames.

\begin{figure*}[h!]
    \begin{minipage}[t]{\linewidth}
	\centering
  \includegraphics[width=1.0\linewidth]{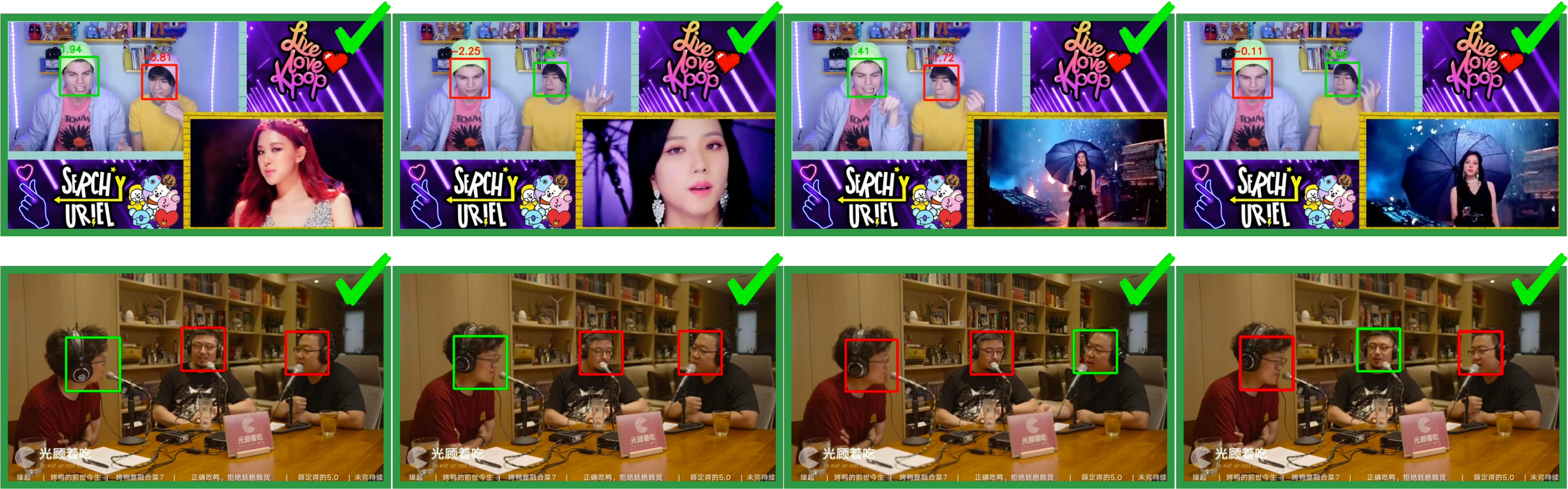}
    \end{minipage}
  \caption{\textbf{Qualitative Results on WASD dataset}. We correctly identifies active speakers across diverse scenarios, including off-the-screen scenarios (top) and podcast-style settings with multiple speakers (bottom). All displayed frames are correctly classified(best viewed zoomed in).}
    \label{fig:supp_wasd}
    \vspace{-3mm}
\end{figure*}

\subsection{Generalization to Other Backbones}
To verify that the proposed consistency losses generalize beyond Light-ASD, we applied $C^3$ to TalkNet and ADENet without any architecture-specific modification. As shown in Tab.\ref{tab:generalization}, consistent improvements are observed across all corruption conditions for both backbones, demonstrating that the proposed losses are architecture-agnostic and can be seamlessly integrated into existing ASD frameworks.
\begin{table}[h]
\vspace{-3mm}
\centering
\caption{\textbf{Generalization across backbones (mAP \%)}.}
\label{tab:generalization}
\resizebox{\linewidth}{!}{%
\begin{tabular}{lcccccccc}
\toprule
\multirow{2}{*}{Method} & Audio & Audio & Visual & MUSAN + & MUSAN + & DEMAND + & DEMAND + \\
& (MUSAN) & (DEMAND) & (Basic) & Obj. Occ. & Pix. & Obj. Occ. & Pix. \\
\midrule
TalkNet      & 88.47 & 88.25 & 81.52 & 62.41 & 88.25 & 62.62 & 88.00 \\
TalkNet+$C^3$                       & \textbf{88.66} & \textbf{88.52} & \textbf{84.44} & \textbf{68.29} & \textbf{88.54} & \textbf{68.14 }& \textbf{88.01} \\
\midrule
ADENet        & 84.58 & 86.50 & 77.70 & 59.81 & 84.73 & 61.07 & 85.12 \\
ADENet+$C^3$                        &  \textbf{85.05}  &  \textbf{86.63}  & \textbf{81.31}  & \textbf{61.88} & \textbf{85.98} &   \textbf{61.88} & \textbf{86.41} \\
\bottomrule
\end{tabular}}
\vspace{-3mm}
\end{table}

\section{Ablation Study under Corruptions}
Tab.\ref{tab:ablation_corrupted} confirms that all three losses are complementary: each individual loss provides incremental gains, and their combination achieves the best average mAP.

\begin{table}[t]
\vspace{-7mm}
\centering
\caption{\textbf{Ablation under joint audio-visual corruption (mAP)}.}
\label{tab:ablation_corrupted}
\resizebox{\linewidth}{!}{%
\begin{tabular}{ccc|cccccccc|c}
\toprule
Inter & Intra & Pred & Park & Cafe & Metro & River & Rest. & Cafeter. & Pub.Sta. & Meet. & Avg \\
\midrule
\xmark & \xmark & \xmark & 85.09 & 80.50 & 85.86 & 83.62 & 79.94 & 80.50 & 83.83 & 78.85 & 82.27 \\
\cmark & \xmark & \xmark & 91.57 & 88.53 & 92.00 & 90.20 & 88.29 & 88.53 & 90.61 & 87.63 & 89.67 \\
\xmark & \cmark & \xmark & 91.78 & 88.54 & \textbf{92.31} & 90.79 & 88.26 & 88.54 & 90.89 & 87.67 & 89.85 \\
\xmark & \xmark & \cmark & 91.66 & 88.68 & 92.01 & 90.61 & 88.42 & 88.68 & 90.67 & 87.69 & 89.96 \\
\cmark & \cmark & \xmark & 91.37 & 88.13 & 91.75 & 90.43 & 88.17 & 88.13 & 90.51 & 87.36 & 89.48 \\
\cmark & \xmark & \cmark & 92.07 & 89.08 & 92.29 & 90.67 & 88.62 & 89.08 & 90.78 & 88.14 & 90.09 \\
\xmark & \cmark & \cmark & 91.66 & 88.68 & 92.01 & 90.61 & 88.42 & 88.68 & 90.67 & 87.69 & 89.96 \\
\cmark & \cmark & \cmark & \textbf{92.00} & \textbf{89.11} & 92.28 & \textbf{90.79} & \textbf{88.91} & \textbf{89.11} & \textbf{91.03} & \textbf{88.30} & \textbf{90.19} \\
\bottomrule
\end{tabular}}
\vspace{-5mm}
\end{table}

\section{Hyperparameter Sensitivity}
Tab.~\ref{tab:hparam-sweep} shows results across a range of hyperparameter settings, where the selected configuration (bold) achieves the best performance while remaining stable across neighboring values, suggesting the method is not highly sensitive to these choices.

\begin{table}[t]
\vspace{-7mm}
\centering
\caption{\textbf{Hyperparameter sweep on AVA-ActiveSpeaker}.}
\label{tab:hparam-sweep}
\setlength{\tabcolsep}{20pt}
\renewcommand{\arraystretch}{0.8}
\small
\begin{tabular}{ccccc}
\toprule
$\lambda_{\text{inter}}$ & $\lambda_{\text{intra}}^{A}$ & $\lambda_{\text{intra}}^{V}$ & $\lambda_{\text{pred}}$& mAP (\%) \\
\midrule
0.001 & 0.0001 & 0.0001 & 0.001 & 93.51 \\
0.001 & 0.001 & 0.001 & 0.001   & 93.48 \\
\textbf{0.01}  & \textbf{0.001}  & \textbf{0.001}  & \textbf{0.01}  & \textbf{93.83} \\
0.01  & 0.01  & 0.01  & 0.01  & 93.40 \\
0.1  & 0.01  & 0.01  & 0.1  & 93.42 \\
0.1  & 0.1  & 0.1  & 0.1  & 93.34 \\

\bottomrule
\end{tabular}
\end{table}

\end{document}